%% file: egpaper_final.tex
\documentclass[10pt,twocolumn,letterpaper]{article}

\usepackage{iccv}
\usepackage{times}
\usepackage{epsfig}
\usepackage{graphicx}
\usepackage{amsmath}
\usepackage{amssymb}
\usepackage[dvipsnames]{xcolor}
\usepackage[accsupp]{axessibility}
\usepackage{multirow}
\usepackage{booktabs}
\usepackage{algorithm}
\usepackage{algpseudocode}
\usepackage{xr}
\usepackage{caption}
\usepackage{subcaption}
\usepackage[pagebackref=true,breaklinks=true,letterpaper=true,colorlinks,bookmarks=false]{hyperref}

\usepackage[capitalize]{cleveref}
\usepackage{fancyhdr}

\usepackage[breaklinks=true,bookmarks=false]{hyperref}

\iccvfinalcopy 

\def\iccvPaperID{10121} 

\ificcvfinal\fi

\begin{document}

\title{What do neural networks learn in image classification? \\ A frequency shortcut perspective }

\author{Shunxin Wang
\and
Raymond Veldhuis
\and 
Christoph Brune
\and
Nicola Strisciuglio\and
University of Twente, The Netherlands
}
\maketitle

\begin{abstract}
    Frequency analysis is useful for understanding the mechanisms of representation learning in neural networks (NNs). Most research in this area focuses on the learning dynamics of NNs for regression tasks, while little for classification. This study empirically investigates the latter and expands the understanding of frequency shortcuts. First, we perform experiments on synthetic datasets, designed to have a bias in different frequency bands.  Our results demonstrate that NNs tend to find simple solutions for classification, and what they learn first during training depends on the most distinctive frequency characteristics, which can be either low- or high-frequencies. Second, we confirm this phenomenon on natural images. We propose a metric to measure class-wise frequency characteristics and a method to identify frequency shortcuts. The results show that frequency shortcuts can be texture-based or shape-based, depending on what best simplifies the objective. Third, we validate the transferability of frequency shortcuts on out-of-distribution (OOD) test sets. Our results suggest that frequency shortcuts can be transferred across datasets and cannot be fully avoided by larger model capacity and data augmentation. We recommend that future research should focus on effective training schemes mitigating frequency shortcut learning. Codes and data are available at \url{https://github.com/nis-research/nn-frequency-shortcuts}.
\end{abstract}

\section{Introduction}
\label{sec:intro}

 Deep neural networks (DNNs) have been widely used to tackle problems in many fields, e.g. medical data analysis, self-driving vehicles, robotics, and surveillance. However, the underlying predictive processes of DNNs are not completely understood due to the black-box nature of their nonlinear multilayer structure~\cite{12116}. While a DNN can approximate any function~\cite{pmlr-v97-rahaman19a}, its (hundreds of) millions of parameters limit the understanding of function approximation process. 
 Analyzing the learned features is a viable way to understand what triggers the predictions, although explaining how DNNs process data needs further exploration~\cite{Tjoa_2021}.
 
 Researchers worked on explaining the predictions of NNs in terms of their input, using Saliency~\cite{6034}, Gradient-weighted Class Activation Mapping~\cite{Selvaraju_2019} and Layer-wise Relevance Propagation~\cite{0130140}. These techniques highlight the area of an image that contributes to prediction but do not explain why the performance of NNs degrades on OOD data. 
 \begin{figure}[!t]
    \centering
    \includegraphics[width=0.95\linewidth]{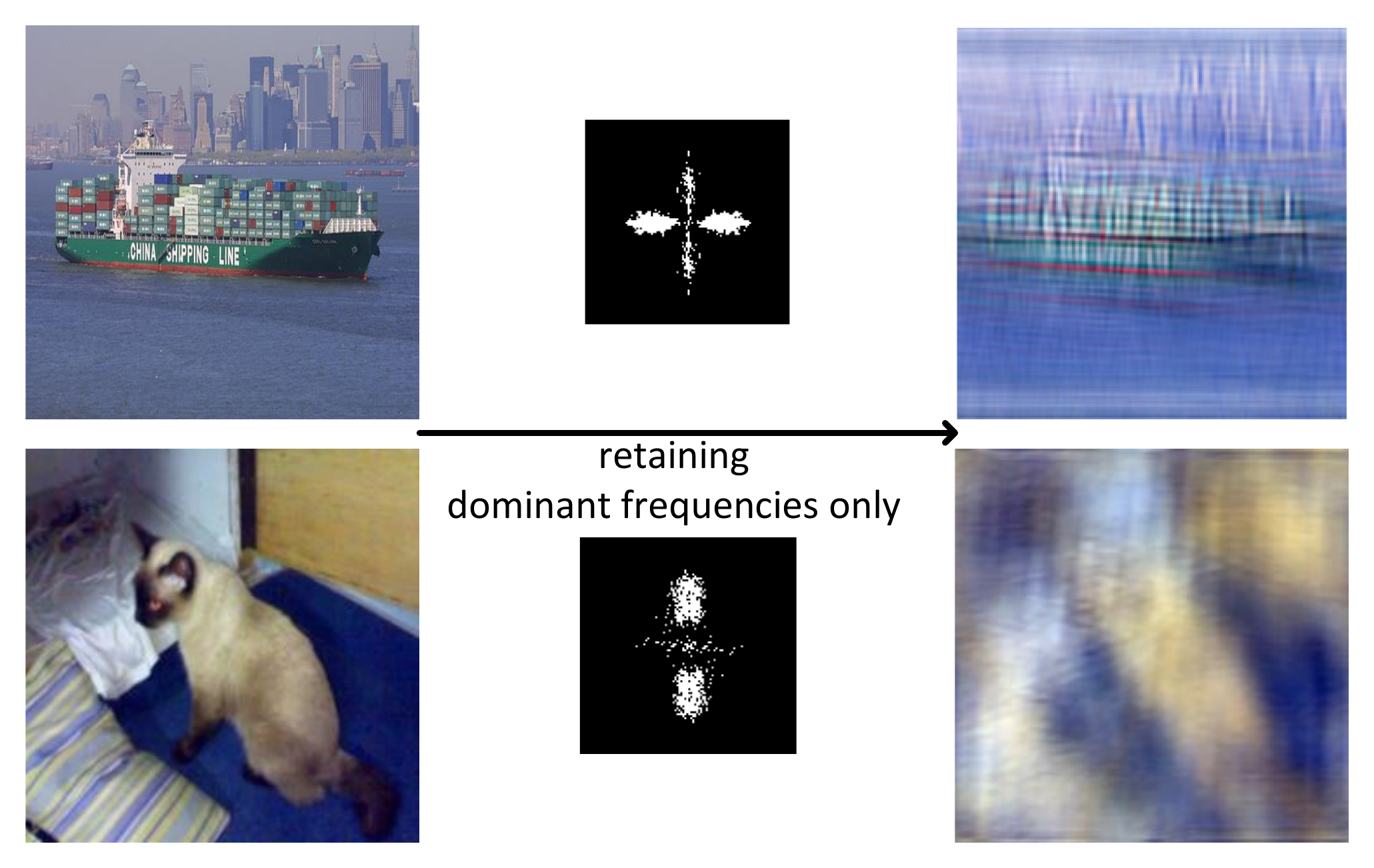}
    \caption{Images of `container ship' and 'siamese cat' and their DFM-filtered versions with only top-$5\%$ dominant frequencies  (the white dots in the central figures) retained can both be recognized correctly by NNs.}
    \label{fig:dfmfilter}
\end{figure}
 Recently, an interest in understanding the learning dynamics of NNs from a frequency perspective has grown. NNs are found to learn lower frequencies first in regression tasks~\cite{pmlr-v97-rahaman19a}, as they carry most of the needed information to reconstruct signals~\cite{14313}. Thus NNs tend to fit low-frequency functions first to data~\cite{CiCP-28-1746}. This biased learning behavior is known as simplicity bias~\cite{Shah2020}, which induces the NNs to learn simple but effective patterns, i.e. shortcuts solutions that disregard semantics related to the problem at hand but are simpler for solving the optimization task. For instance, the frequency shortcuts proposed in~\cite{wang2022frequency} are sets of frequencies used specifically to classify certain classes. 

In this work, we empirically analyze the learning dynamics of NNs for image classification and relate it to simplicity-bias and shortcut learning from a frequency perspective.
Our results indicate that simplicity-biased learning in NNs leads to frequency-biased learning, where the NNs exploit specific frequency sets, namely \emph{frequency shortcuts}, to facilitate predictions.  These frequency shortcuts are data-dependent and can be either texture-based or shape-based, depending on what best simplifies the objective function (e.g. a unique color, texture, or shape associated with a particular class in a dataset, without necessarily other meaningful semantics). This may impact generalization. We demonstrate this phenomenon through texture-based and shape-based frequency shortcuts in~\cref{fig:dfmfilter}. When we retain only specific subsets of frequencies (identified using a method proposed in this paper) from images of `container ship' and `siamese cat', the classifier can recognize them correctly. Interestingly, when the same sets of frequencies are retained from images of other classes, the predictions are biased towards these two classes, indicating that the frequency sets are specific for their classification.  

Different from previous work on regression tasks~\cite{pmlr-v97-rahaman19a}, we investigate the learning dynamics and frequency shortcuts in NNs for image classification. Compared to the work uncovering frequency shortcuts~\cite{wang2022frequency}, we expand the understanding of them and demonstrate that they can be texture, shape, or color, depending on data characteristics. We propose a metric to compare the frequency characteristics of data and investigate systematically the impact of present/absent shortcut features on OOD generalization.
In summary, our \textbf{contributions} are: 
\begin{enumerate}
\item We complement existing studies that showed NNs for regression tasks are biased towards low-frequency~\cite{pmlr-v97-rahaman19a}. For classification, we find that NNs can exhibit different frequency biases, tending to adopt frequency shortcuts based on data characteristics because of simplicity-bias learning. Our analysis provides valuable insights into the learning dynamics of NNs and the factors influencing their behavior.
\item  We propose a method to identify frequency shortcuts, based on culling frequencies that contribute less to classification.
These shortcuts are composed of specific frequency subsets that correspond to textures, shapes, or colors, providing further insight into the texture-bias identified by Geirhos~\etal~\cite{geirhos2018imagenettrained} and background-dependency found in~\cite{200609994}. 

\item  We systematically examine the influence of frequency shortcuts on the generalization of NNs and find that the presence of frequency shortcut features in an OOD test set may give an illusion of improved generalization. Furthermore, we find that larger model capacity and common data augmentation techniques like AutoAugment~\cite{Cubuk_2019_CVPR}, AugMix~\cite{hendrycks2020augmix}, and SIN~\cite{12231} cannot fully avoid shortcut learning. We recommend further research targeting frequency information to avoid frequency shortcut learning. 
\end{enumerate}

\section{Related works}
\label{relatedwork}
\paragraph{Frequency analysis.}
Recently, Fourier interpretations of NNs were published.  For regression tasks, NNs tend to learn low-frequency components first~\cite{pmlr-v97-rahaman19a, Xu_2020}, while initial layers bias towards high-frequency components~\cite{03972}. In classification, NNs exhibit a bias towards middle-high frequency during testing~\cite{Abello_2021_CVPR}. The authors in~\cite{Abello_2021_CVPR} argued that the importance of frequency is data-driven. Sensitivity to different frequency perturbations was measured in~\cite{ NEURIPS2019_b05b57f6}, showing that most NNs are more sensitive to middle-high frequency noise. The impact of high-frequency dependence on the robustness of NNs was investigated in~\cite{9156428}. These analyses show that NNs for regression and classification tasks exhibit different frequency dependencies, while there is a lack of analysis on the learning dynamics of NNs for classification. We study what and how NNs learn in classification, highlighting their data-driven behavior and complementing existing work on regression tasks. We uncover that NNs can learn to use specific frequency sets encompassing both low and high frequencies to achieve accurate classification. 

\paragraph{Shortcut learning.} 

In classification, decision rules based on spurious correlations between data and ground truth, rather than semantic cues, are known as shortcuts~\cite{Geirhos_2020}. For example, a network may classify images based on the presence of text embedded in the images, rather than the actual image content~\cite{lapuschkin2019unmasking}, negatively impacting generalization~\cite{wang2023robustness}.
Identifying shortcuts learned by NNs might be helpful to avoid unwanted learning behavior and thus improve generalization. It is easy to identify shortcuts that are artificially added and are visible (e.g. color patches~\cite{diagnostics12010040}, line artefacts~\cite{06406}, or added text~\cite{lapuschkin2019unmasking}). However, for those implicitly existing in data (e.g. particular textures or shapes), their identification is difficult. Most methods focus on mitigating learning shortcut information in data~\cite{06922,08822,pezeshki2021gradient,11230}, rather than explicitly identifying them. Wang~\etal~\cite{wang2022frequency} investigated shortcut learning from a frequency perspective and proposed the definition of frequency shortcuts. However, their algorithm for shortcut identification is heavily influenced by the order of frequency removal and their observations are limited to texture-based shortcuts. In this paper, our frequency shortcut identification method does not have such limitations. We broaden the understanding of frequency shortcuts, study the data-dependency of shortcut features,
and provide a more systematic analysis of the impact of shortcuts on OOD generalization. 



\section{Frequency shortcuts in image classification}
\label{sec:learningbeh}
For regression tasks, it is known that NNs are biased towards learning low-frequency components (LFCs) first during training~\cite{pmlr-v97-rahaman19a}. This has not been verified for classification tasks. Here we study the learning behavior of NNs in image classification and its relation to shortcut learning and simplicity-bias, using both synthetic (\cref{section:synthetic}) and natural images (\cref{sec:natural}).
We use synthetic data to study the learning behavior of NNs and show their tendency to discover shortcuts in the frequency domain. Inspired by the insights gained on the synthetic data, we propose a method based on frequency culling to examine the frequency dependency of NNs trained on natural images,  which contain intricate frequency information. This allows us to uncover the frequency shortcuts learned by NNs for classification.
 

\subsection{Experiments on synthetic data}
\label{section:synthetic} 
\begin{figure}[!t]
     \centering
     \includegraphics[width = 0.95\linewidth]{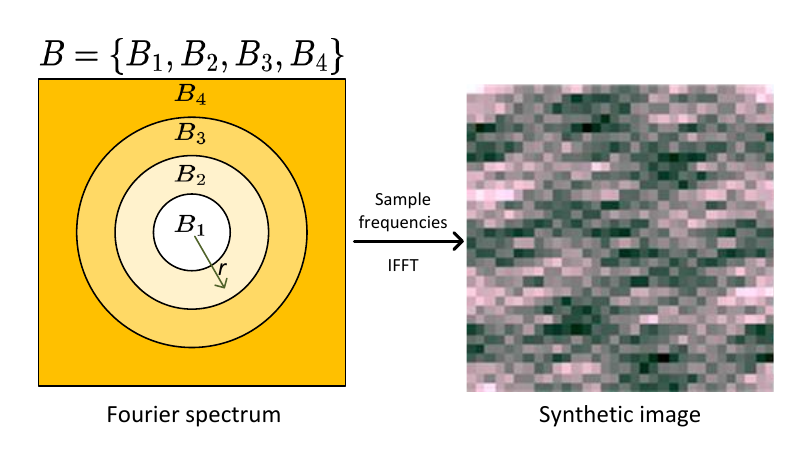}
     \vspace{-5mm}
     \caption{Evenly separated frequency bands. $B_1$ denotes the lowest band and $B_4$ denotes the highest  one.}
     \label{fig:band}
\end{figure}

\paragraph{Design of synthetic datasets.}
To study the impact of data characteristics on the spectral bias of NNs and frequency shortcut learning,  we generate four synthetic datasets, each with a frequency bias in a different band, from low to high. This allows us to examine the effect of different frequency biases on the learning behavior of NNs. 
We separate evenly the Fourier spectrum into four frequency bands (see~\cref{fig:band}). The bands are denoted by $B_1$ the lowest frequency band, $B_2$ and $B_3$ the mid-frequency bands, and $B_4$ the highest frequency band. 
Each dataset contains four classes and images of $32 \times 32$ pixels. 
An image is generated by sampling at least eight frequencies from the frequency bands associated with the target class (see~\cref{tab:detailsyn}), according to a probability density function:
\begin{equation*}
Pr(r) = S \cdot \frac{1}{r+1},    
\textrm{~~~~with }
S = \frac{1}{\sum_{r=1}^{R}\frac{1}{r+1}}.
\end{equation*}
\noindent $R$ is the largest radius and $r=\sqrt{u^2+v^2}$ is the radius of frequency $[u,v]$. This prioritizes the sampling of LFCs, mimicking the frequency distribution of natural images.

We use $b \in B = \{B_1, B_2, B_3, B_4\}$ to control the frequency bias in the generated data. For instance, in the dataset Syn$_{b}$ with $b = B_1$, the frequency bands for classes $C_0$ and $C_1$ are $\{B_2, B_3, B_4\}$ while class $C_3$ has frequency band $B_1$. To distinguish between $C_0$ and $C_1$, we embed \emph{special patterns} consisting of a set of frequencies $[u,v]$ ($u=v\in \{1, 3, 5, 7, 9, 11, 13, 15\}$) into the images of class $C_0$ which are removed from the images of other classes.  The design imposes various levels of classification difficulty by incorporating different levels of data complexity for each class ($C_3 < C_0 < C_1\approx  C_2$), as observed visually. This aids in comprehending the connection between simplicity-bias learning and spectral-bias of NNs in classification.

\paragraph{Hypothesis.}
  As noted in the theory of simplicity-bias~\cite{Shah2020}, NNs tend to achieve their objective in the simplest way. As a result, NNs  for regression tasks approximate LFCs first compared to HFCs~\cite{pmlr-v97-rahaman19a,Xu_2020,11561,Abello_2021_CVPR}.   
Based on this, we hypothesize that NNs might prioritize learning to distinguish classes with the most discriminative frequency characteristics in classification. Thus, what the NNs first learn could depend on data bias rather than being limited to low frequencies. This learning behavior could result in frequency shortcut learning, where the NNs focus on specific frequencies to achieve their objective in a simpler way.

\begin{table}
  \caption{Design details of a synthetic dataset Syn$_{b}$ with $b \in B=\{B_1, B_2, B_3, B_4\}$. The special pattern contains frequencies $[u,v]$ where $u=v\in \{1, 3, 5, 7, 9, 11, 13, 15\}$ are removed from classes other than $C_0$. 
  }
  \label{tab:detailsyn}
  \centering
  \renewcommand{\arraystretch}{0.95}
  \footnotesize
  \begin{tabular}{ccc}
    \toprule
    \bfseries class &  \bfseries frequency bands &  \bfseries special patterns \\
    \midrule
     \bfseries $C_0$ & $B - b$ &   \checkmark  \\
     \bfseries $C_1$ & $B - b$ & - \\
      \bfseries $C_2$ & $B$ & - \\
      \bfseries $C_3$ & $b$ & - \\
    \bottomrule
  \end{tabular}
\end{table}

\begin{figure}
    \centering
    \includegraphics[width = \linewidth]{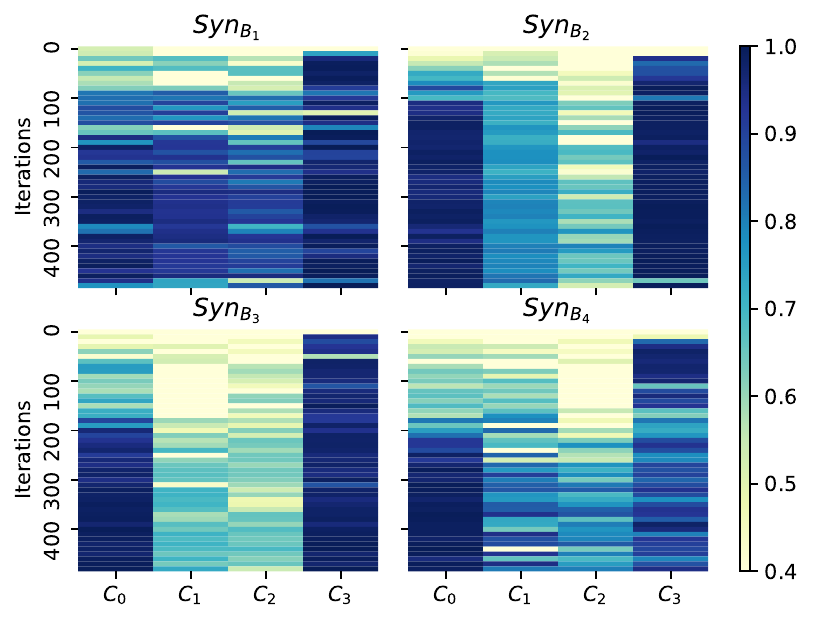}
    \caption{{$F_1$-scores of each class in the first $500$ training iterations. $C_3$ has higher $F_1$-scores than others at the early training stage, meaning that it is learned first even if it only has frequencies sampled from the highest frequency band. }}
    \label{fig:f1scorealexsyn}
\end{figure}

\paragraph{Data characteristics influence what NNs learn first.}
We conduct experiments on the synthetic data to test this hypothesis.  We train ResNet18 models on the synthetic datasets and expect they can distinguish classes like $C_0$ and $C_3$ easily and from the early stages of training, as they carry more distinctive characteristics than others. To evaluate this, we measure their classification performance in the first $500$ iterations of training by computing the F$_1$-score per class. This provides insight into whether each class is correctly classified and how many false positives each class attracts. 
 We report the obtained $F_1$-scores (see~\cref{fig:f1scorealexsyn}) and observe that for class $C_3$ (with a clear frequency bias), the $F_1$-score is generally higher than other classes in the first few iterations, indicating that it is immediately distinguished from others across the four synthetic datasets, followed by class $C_0$.
 This finding suggests that the more distinguishable characteristics of class $C_3$ play an important role in driving the learning behavior of NNs. 
  Note that, despite the bias in different bands across the four synthetic datasets, class $C_3$ is always learned first, indicating that NNs can learn either low- or high-frequency early in training if they are more discriminative than other frequencies. Thus, \textit{what frequencies are learned first by NNs in classification is driven by simplicity-bias and data characteristics.}

\begin{figure}
    \centering
    \includegraphics[width=0.9\linewidth]{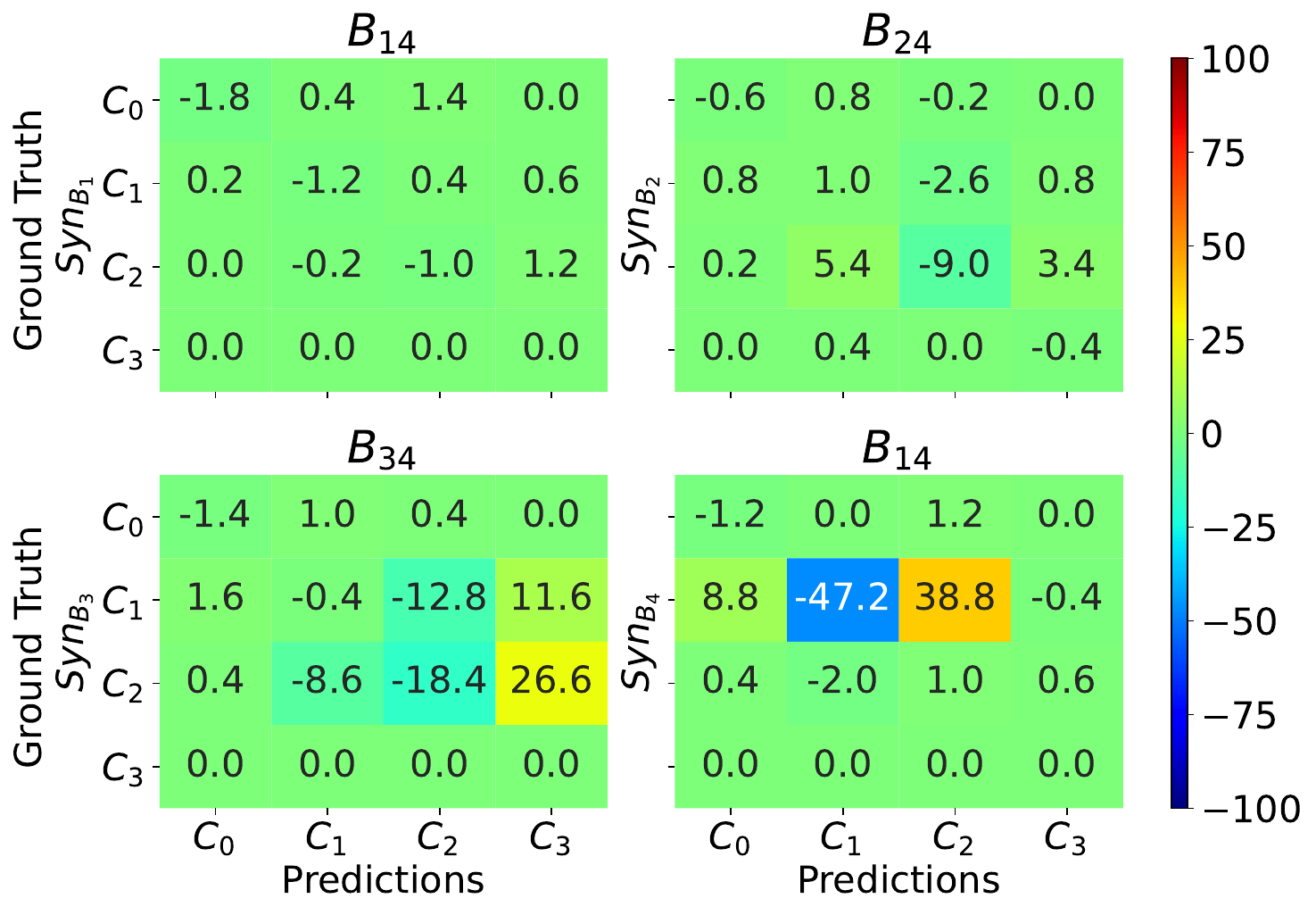}
    \caption{{Relative confusion matrices of models tested on different band-stop synthetic datasets (e.g. $B_{14}$ indicates the bands $B_1$ and $B_4$ are used). The top-left figure shows the comparison of the results on the original test set and its band-stopped version for the model trained on $Syn_{B_1}$. Other matrices show the results of other models. Most $\Delta^{C_i, C_i}$ ($i = 0,1,2,3$) values are close to or larger than 0, indicating good performance on band-stopped datasets due to learned frequency shortcuts.  }}
    \label{fig:C1s_par}
\end{figure}

\paragraph{Data bias and simplicity bias can lead to frequency shortcuts.}
Based on the frequency characteristics of the synthetic datasets, we examine how NNs find shortcuts in the Fourier domain by comparing the classification results of the NNs tested on the original synthetic datasets and their band-stop versions where two frequency bands in $B$ are removed. We report the results using relative confusion matrices (see~\cref{fig:C1s_par}),  computed as: 
\begin{equation*}
  \Delta^{C_i,C_j} = (Pred_{bs}^{C_i,C_j}-Pred_{org}^{C_i,C_j})/N_c\times 100, 
\end{equation*}
 \noindent where $Pred_{bs}^{C_i,C_j}$ is the number of samples from class $C_i$ in the band-stopped test set predicted as class $C_j$, $Pred_{org}^{C_i,C_j}$ is the equivalent on the original test set, and $N_C$ is the number of samples in class $C_i$. 
 
 As $\Delta^{C_i, C_i}$ ($i = 0,1,2,3$) is larger than or equal to zero, the performance of the model improves or remains the same on the band-stop test sets, indicating that the limited bands provide enough discriminative information for classification, while negative values indicate lower performance.  Class $C_2$ in the four synthetic datasets is designed to contain frequencies from all bands. If a model can predict class $C_2$ using only frequencies from partial bands instead of considering frequencies across the whole spectrum, then it is considered to likely be using frequency shortcuts to classify  $C_2$. Observed from~\cref{fig:C1s_par}, $\Delta^{C_2, C_2}$ are -1 and 1 for models trained on $Syn_{B_1}$ and $Syn_{B_4}$ respectively. The good performance indicates that NNs apply frequency shortcuts in the limited bands for classifying samples of $C_2$.  
 Moreover,  $\Delta^{C_0, C_0}$ of models trained on the four synthetic datasets are close to 0, demonstrating that the NNs can recognize samples of $C_0$ when only part of the frequencies (shortcuts) associated with the \emph{special patterns} are present in the test data. Similar behaviors are observed for other architectures (see results of AlexNet and VGG  in the supplementary material). To summarize,  the NNs trained on the four synthetic datasets use frequency differently, but they all adopt frequency shortcuts depending on the data characteristics.

\subsection{Experiments on natural images}
\label{sec:natural}
The synthetic experiments show frequency characteristics of data affect what NNs learn. To analyze the more intricate frequency distributions of natural images, we introduce a metric to compare the average frequency distributions of individual classes within a dataset. This facilitates the identification of discriminative and simple class-specific frequency characteristics to learn early in training. While this metric provides valuable insights into the potential learning behavior, a deeper examination of frequency usage by NNs is also needed. To this end, we propose a technique based on frequency culling, which can help uncover frequency shortcuts explicitly. Additionally, we investigate how model capacity and data augmentation impact shortcut learning. As NNs are found to exhibit texture-bias~\cite{geirhos2018imagenettrained} on natural images, we specifically augment data using SIN to create a dataset with more shape-bias. This better demonstrates how texture-/shape-biased data characteristics affect frequency shortcut learning.

\paragraph{A frequency distribution comparison metric.}

From the insights gained on the synthetic experiments, we recognize the importance to examine the frequency characteristics of individual classes within a dataset to understand comprehensively what NNs learn. Thus, we devise a metric called Accumulative Difference of Class-wise average Spectrum (ADCS), which considers that NNs are amplitude-dependent for classification~\cite{Chen_2021_ICCV}. We compute the average amplitude spectrum difference per channel for each class within a set $C=\{c_0, c_1, \dots, c_n\}$ and average it into a one-channel ADCS. The ADCS for class $c_i$ at a frequency $(u,v)$ is calculated as:

\begin{equation*}
    ADCS^{c_i}(u,v) = \sum_{\substack{\forall c_j \in C \\ c_j \neq c_i}} sign(E_{c_i}(u,v) - E_{c_j}(u,v)),
\end{equation*}
\noindent where
\begin{equation*}
    E_{c_i}(u,v) = \frac{1}{\lvert X^i \lvert} \sum_{x \in X^i} \lvert \mathcal{F}_x(u,v) \lvert
\end{equation*}
\noindent is the average Fourier spectrum for class $c_i$, $x$ is an image from the set $X^i$ of images contained in that class, and $\mathcal{F}_x(u,v)$ is its Fourier transform. $ADCS^{c_i}(u,v)$ ranges from $1-|C|$ to $|C|-1$. A higher value indicates that a certain class has more energy at a specific frequency than other classes.

\paragraph{Impact of class-wise frequency distribution on the learning process of NNs.}
We choose ImageNet-10~\cite{huang2021unlearnable}, a reduced version of ImageNet~\cite{5206848} for the following analysis.  It has lower computational requirements and greater manageability, compared to the full ImageNet dataset. For larger datasets with more classes, one may expect severer shortcut learning behaviors, as the NNs will tend to find quick solutions to simplify a more difficult classification problem. 

\begin{figure}
    \centering    
    \begin{subfigure}[b]{0.23\textwidth}
         \centering
         \includegraphics[height=2.8cm]{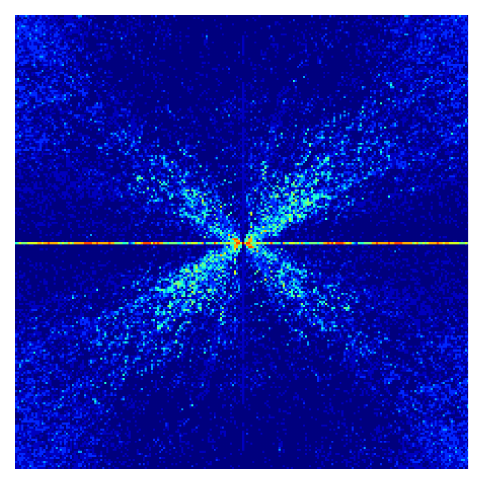}
         \caption{$ADCS^{humming\ bird}$}
         \label{fig:acc_olp2}
     \end{subfigure}%
     \begin{subfigure}[b]{0.23\textwidth}
         \centering
         \includegraphics[height=2.8cm]{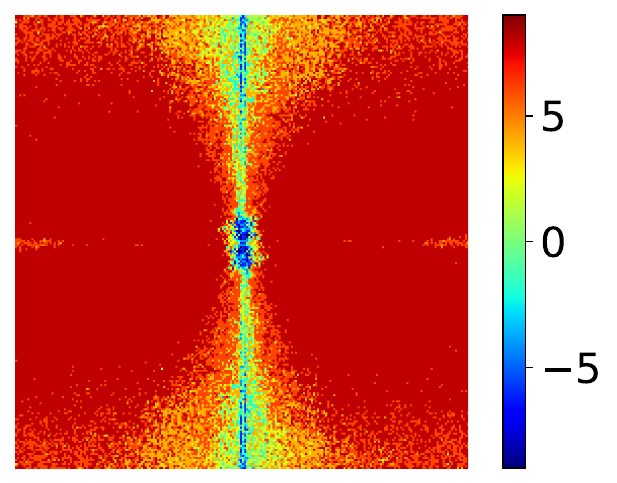}
         \caption{$ADCS^{zebra}$}
         \label{fig:acc_olp7}
     \end{subfigure}
\caption{ADCS of classes `humming bird' and `zebra'.}
    \label{fig:ADCS}
\end{figure}

Using ADCS, we find that the classes `humming bird' and `zebra' possess certain distinctive frequency characteristics that can be readily exploited by models to distinguish them from other classes at early training stages.  The resulting ADCS of `\emph{humming bird}' (see~\cref{fig:acc_olp2}) indicates that samples from this class have on average much less energy than other classes across almost the whole spectrum. Conversely, the ADCS of `\emph{zebra}' (see~\cref{fig:acc_olp7})  reveals that images from this class have a marked energy preponderance in the middle and high frequencies, as indicated by the prominence of red color in these frequency ranges. 

\begin{figure}
    \centering
     \begin{subfigure}[b]{0.5\textwidth}
        \centering
        \includegraphics[width = \textwidth]{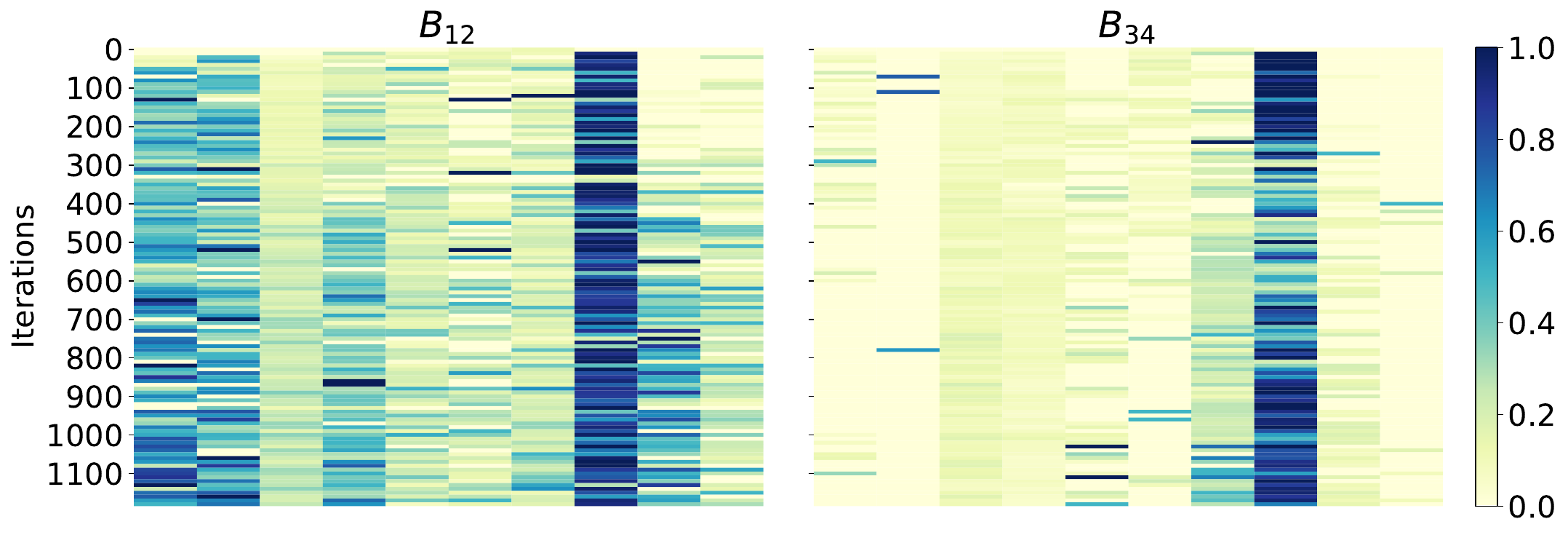}
        \caption{Precision of each class in the first $1200$ iterations.}
        \label{fig:learimagenet10Precision}
    \end{subfigure}
    \begin{subfigure}[b]{0.5\textwidth}
        \centering
        \includegraphics[width = \textwidth]{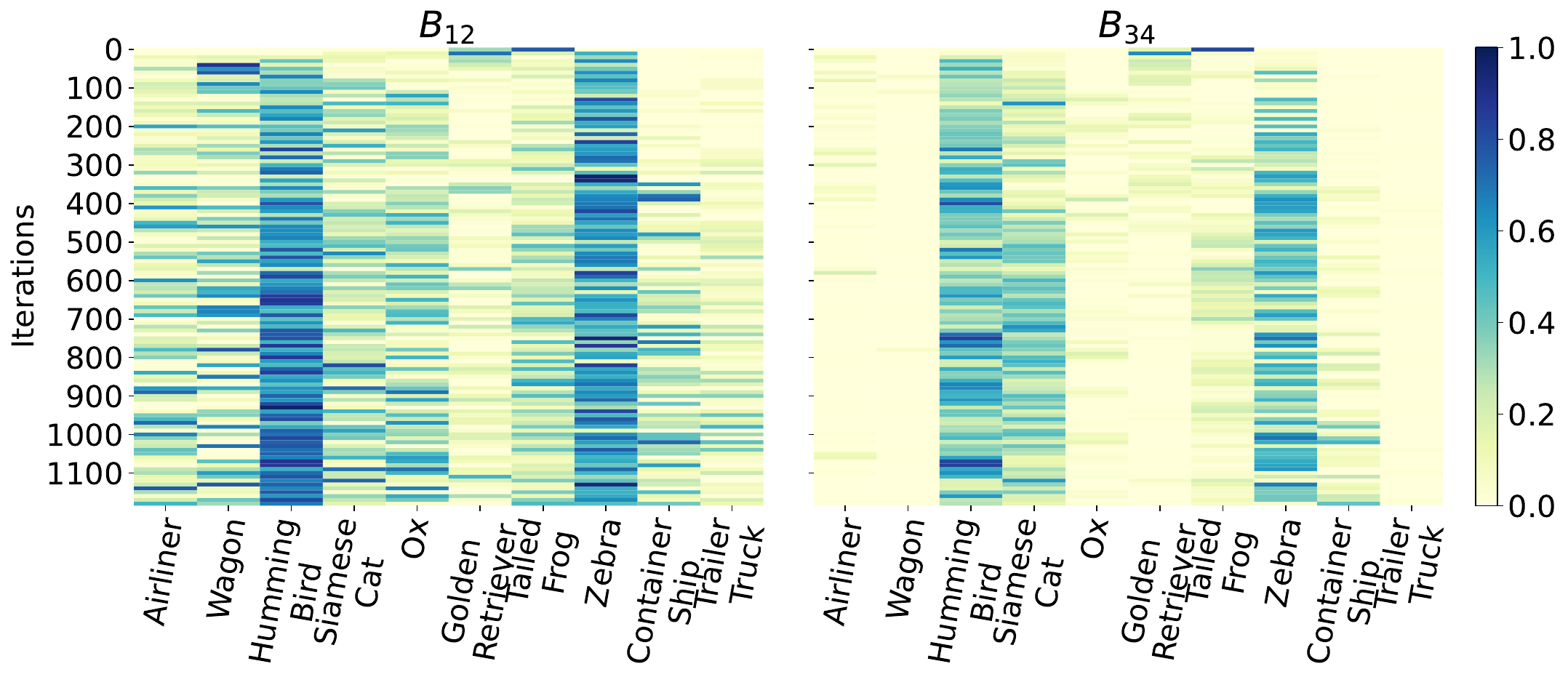}
        \caption{Recall of each class in the first $1200$ iterations).}
        \label{fig:learimagenet10Recall}
    \end{subfigure}     
    \caption{{Precision and recall rates of ResNet18 trained on ImageNet-10 for the first $1200$ iterations.}}
    \label{fig:result_imagenet10}
\end{figure}

To verify the impact of such frequency characteristics on the learning behavior, we train NNs on ImageNet-10.
We inspect the frequency bias in the early training phase, by testing models on low- and high-pass versions of the dataset for the first  $1200$ training iterations, rather than the original test set.
We compute the recall and precision of each class and observe that the precision of class `\textit{zebra}' (see~\cref{fig:learimagenet10Precision}) and the recall of class `\textit{humming bird}' (see~\cref{fig:learimagenet10Recall}) are generally higher than those of other classes. This shows that these two classes are learned faster than others. 
In summary,  our findings indicate that NNs for classification can learn and exploit substantial spectrum differences among classes, which serve as highly discriminative features at the early learning stage. This further supports our previous observations in synthetic datasets that\textit{ what is learned first by NNs is influenced by the frequency characteristics of data}.

\begin{figure*}[!t]
    \centering
    \includegraphics[width=\linewidth]{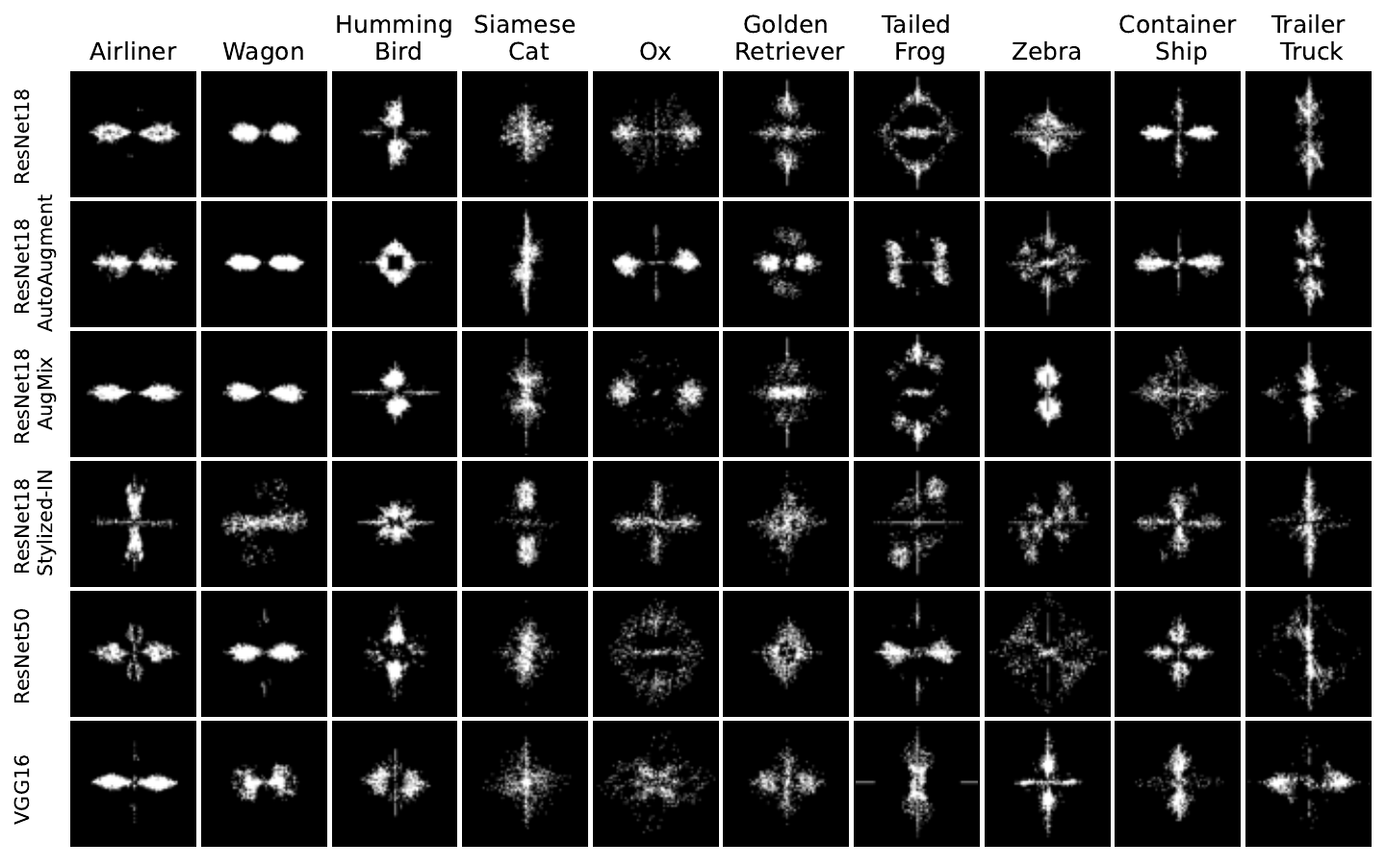}
    \caption{Dominant frequency maps of ResNet18 (with AutoAugment/AugMix/SIN), ResNet50 and VGG16. The maps show the top-$5\%$ dominant frequencies of each class in ImageNet-10. }
    \label{fig:masksameimage10}
\end{figure*}

\paragraph{A frequency shortcut identification method.}

To identify frequency shortcuts, we propose a method based on culling irrelevant frequencies, similar to the analysis strategy in~\cite{Abello_2021_CVPR}. We measure the relevance of each frequency to classification by recording the change in loss value when testing a model on images of a certain class with the concerned frequency removed from all channels. The increment in loss value is used as a score to rank the importance of frequencies for classification. Frequencies with higher scores are considered more relevant for classification, as their absence causes a large increase in loss.
We compute a one-channel \textbf{dominant frequency map (DFM)} for a class by selecting the top-$X\%$ frequencies according to the given ranking. Using the DFMs, we study the effect of dominant frequencies on image classification and the extent to which they indicate frequency shortcuts (specific sets of frequencies leading to biased predictions for certain classes). 
To quantify these, we classify all images in the test set retaining only the  top-$X\%$ frequencies of a certain class (i.e. top-$X\%$ DFM-filtered test set). We calculate the true positive rate (TPR) and false positive rate (FPR) to evaluate their discrimination power and specificity for a certain class, respectively. We consider classes with high TPR and FPR as instances where the classifier is induced to learn and apply frequency shortcuts. 

\paragraph{Frequency shortcuts can be texture- or shape-based.}
We show the DFMs with the top-$5\%$ frequencies for ResNet(s) trained w/o or w/ augmentation (AutoAugment, AugMix, and SIN) and VGG16 in~\cref{fig:masksameimage10} (more DFMs are in the supplementary material). In~\cref{tab:noFrqs}, we report the TPR and FPR of models tested on the original and the top-$5\%$ DFM-filtered test sets. For ResNet18, the TPR and FPR of classes `zebra' and `container ship' are higher than other classes, indicating that the model applies frequency shortcuts for these two classes. Similarly, for ResNet18 trained with SIN which replaces object textures to emphasize shape information, the model learns a frequency shortcut for class `siamese cat'. In~\cref{fig:dfmfilter}, we show examples of `container ship' and `siamese cat' images, their corresponding DFMs, and the images retaining only the frequencies in the DFMs, which contain textures, shapes, or colors that would not be used alone by human observers to classify images, but that NNs can exploit solely due to frequency shortcut learning.
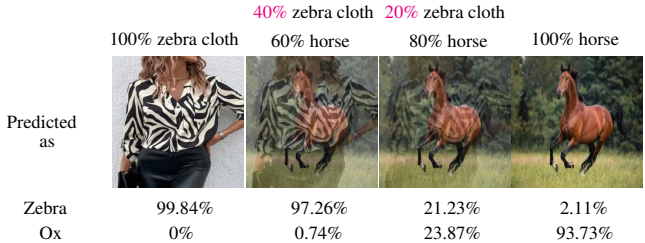
\begin{figure}[!t]
    \centering
    \input{imgs/zebracloth_new.tex}
    \caption{Model classifies zebra-pattern clothes with high confidence but misclassifies horse as ox. Mixing images of `zebra cloth' and `horse' increases the confidence of `zebra' predictions. This indicates that the model relies on texture over shape information, its ability to generalize and recognize another animal of similar shape but different texture.  }
    \label{fig:zebracloth}
\end{figure}
\begin{table*}
\tiny\addtolength{\tabcolsep}{-5pt}
  \caption{ID test: TPRs and FPRs on ImageNet-10 and the top-$5\%$ DFM-filtered versions (w/ df).}
  \label{tab:noFrqs}
  \centering
  \renewcommand{\arraystretch}{0.85}
  \begin{tabular}{p{1.9cm}p{0.9cm}c@{\hspace{16\tabcolsep}}c@{\hspace{8\tabcolsep}}c@{\hspace{8\tabcolsep}}c@{\hspace{8\tabcolsep}}c@{\hspace{8\tabcolsep}}c@{\hspace{8\tabcolsep}}c@{\hspace{8\tabcolsep}}c@{\hspace{8\tabcolsep}}c@{\hspace{8\tabcolsep}}c@{\hspace{8\tabcolsep}}c@{\hspace{8\tabcolsep}}c}
    \toprule
    \multicolumn{13}{c}{\bfseries ImageNet-10} \\ \midrule
      \bfseries Model &  & & \bfseries airliner     & \bfseries wagon  & \bfseries humming bird & \bfseries siamese cat & \bfseries ox & \bfseries golden retriever & \bfseries tailed frog & \bfseries zebra & \bfseries container ship & \bfseries trailer truck & \bfseries average \\
    \midrule
   \bfseries ResNet18 &&$TPR$  & 0.96   & 0.8 & 0.94   & 0.98   & 0.92   & 0.9    & 0.84   & 0.96   & 0.94   & 0.96  &  0.92   \\
     && $FPR$  &0.0044 & 0   & 0.0178 & 0.0067 & 0.0156 & 0.0022 & 0.0044 & 0.0022 & 0.0133 & 0.0222  &\\
      \cmidrule{2-14}
      & w/ df&$TPR$  & 0.08   & 0      & 0.4    & 0.8    & 0.02   & 0.02   & 0.14   & \textbf{0.8}    & \textbf{0.54 }  & 0.06 &   \\
     && $FPR$ & 0.0044 & 0      & 0.02   & 0.0356 & 0.0311 & 0.0044 & 0.0022 & \textbf{0.1178} & \textbf{0.1889} & 0.0022    &    \\
    \midrule
    \bfseries ResNet18+AutoAug &&$TPR$  & 0.92   & 0.76 & 0.88   & 0.92   & 0.96   & 0.84   & 0.66   & 0.94   & 0.94   & 0.8  &  0.862  \\
     && $FPR$  &0.0089 & 0    & 0.0289 & 0.0089 & 0.0267 & 0.0111 & 0.0044 & 0.0067 & 0.0222 & 0.0356 &  \\
     \cmidrule{2-14}
      & w/ df&$TPR$  & 0    & 0      & 0      & 0.22   & 0.04   & 0.02   & 0      &\textbf{ 0.26}   & 0.18   & 0  &    \\
     && $FPR$ &0    & 0      & 0      & 0.0067 & 0.0222 & 0.0111 & 0      & 0.0089 & 0.0622 & 0    &   \\
    \midrule
    \bfseries ResNet18+AugMix &&$TPR$  & 0.92   & 0.86   & 0.96   & 0.98   & 0.92   & 0.88   & 0.72   & 0.96 & 0.92   & 0.92 & 0.904 \\
     && $FPR$  &0.0089 & 0.0022 & 0.0267 & 0.0022 & 0.0222 & 0.0044 & 0.0044 & 0    & 0.0156 & 0.02 &\\
    \cmidrule{2-14}
      & w/ df&$TPR$  & 0.08   & 0      & 0.22   & 0.34   & 0.22   & 0.24   & 0.02   & 0.16   & \textbf{0.88 }  & 0.26 &\\
     && $FPR$ & 0.0067 & 0      & 0.0089 & 0.0267 & 0.1511 & 0.0089 & 0      & 0.0067 & \textbf{0.2444} & 0.0067   & \\
     \midrule
    \bfseries ResNet18+SIN &&$TPR$  & 0.96    &  0.86   &  0.94  & 0.96  & 0.98  & 0.86   & 0.76   &  0.96 & 0.96   & 0.92  &  0.916 \\
     && $FPR$  & 0.0022 & 0.0022 & 0.0178 & 0.0111  & 0.0244  & 0  & 0.0044  & 0.0022   & 0.0133  & 0.0156  &\\
    \cmidrule{2-14}
      & w/ df&$TPR$  & {0.46} &  0 & 0.18 & \textbf{0.98 }& 0.06 & 0.6  & 0 & 0.06  & 0.06  & 0.1 &\\
     && $FPR$ & {0.1267} & 0.0022  & 0.0111 & \textbf{0.5467} & 0.0511 & 0.0822 & 0 & 0.0022 & 0.0622 & 0.0133 & \\
    \midrule
    \bfseries ResNet50 &&$TPR$  & 0.9    & 0.78   & 0.86 & 0.94   & 0.86   & 0.82   & 0.78   & 0.94   & 0.94   & 0.8    &   0.862 \\
     && $FPR$ & 0.0044 & 0.0022 & 0.02 & 0.0044 & 0.0267 & 0.0089 & 0.0111 & 0.0089 & 0.0244 & 0.0422   &  \\
    \cmidrule{2-14}
      & w/ df&$TPR$  & \textbf{0.54} & 0 & 0      & 0.42 & 0      & 0.2    & 0 & 0.16   &\textbf{ 0.7 }   & 0.1    &  \\
     && $FPR$ & \textbf{0.22} & 0 & 0.0022 & 0.04 & 0.0022 & 0.0533 & 0 & 0.0489 & \textbf{0.2289} & 0.0156    &   \\
    \midrule
    \bfseries VGG16 &&$TPR$  & 0.96   & 0.84   & 0.92   & 1      & 0.9    & 0.92   & 0.78   & 0.96   & 0.96   & 0.88 & 0.912    \\
     && $FPR$ & 0.0022 & 0.0022 & 0.0222 & 0.0111 & 0.0133 & 0.0044 & 0.0067 & 0.0022 & 0.0133 & 0.02  &     \\
     \cmidrule{2-14}
      & w/ df&$TPR$  & 0.18   & 0      & 0      & 0.66   & 0.22   & 0.12   & 0.04   & 0.06   &\textbf{ 0.7  }  & 0.22   & \\
     && $FPR$ & 0.0133 & 0      & 0      & 0.0444 & 0.1489 & 0.0267 & 0      & 0.0533 & \textbf{0.42}   & 0.0578     &   \\
    \bottomrule
  \end{tabular}
\end{table*}

\begin{table*}
\tiny
\addtolength{\tabcolsep}{-5pt}
  \caption{Transferability test: TPRs and FPRs of ViT-B on the top-$5\%$ DFM (of ResNet18+SIN)-filtered versions.}
  \label{tab:vitb}
  \centering
  \renewcommand{\arraystretch}{0.85}
  \begin{tabular}{p{1.9cm}p{0.9cm}c@{\hspace{16\tabcolsep}}c@{\hspace{8\tabcolsep}}c@{\hspace{8\tabcolsep}}c@{\hspace{8\tabcolsep}}c@{\hspace{8\tabcolsep}}c@{\hspace{8\tabcolsep}}c@{\hspace{8\tabcolsep}}c@{\hspace{8\tabcolsep}}c@{\hspace{8\tabcolsep}}c@{\hspace{8\tabcolsep}}c}
    \toprule
    \multicolumn{13}{c}{\bfseries ImageNet-10} \\ \midrule
      \bfseries Model &  & & \bfseries airliner     & \bfseries wagon  & \bfseries humming bird & \bfseries siamese cat & \bfseries ox & \bfseries golden retriever & \bfseries tailed frog & \bfseries zebra & \bfseries container ship & \bfseries trailer truck  \\
    \midrule
   \bfseries ViT-B
      & w/ df&$TPR$  &  0.34   & 0.02      &  0.28  & \textbf{0.82}   & 0.44   &0.72    & 0.02   & 0.46   & \textbf{0.92} &  0.6 \\
     && $FPR$ &  0.1933   & 0.0022     &  0.0067  & \textbf{0.22}   &  0.08  & 0.0578   & 0.0133   & 0.0289   & \textbf{0.2467} & 0.0333 \\
    \bottomrule
  \end{tabular}
\end{table*}

\begin{table*}
\tiny\addtolength{\tabcolsep}{-5pt}
  \caption{OOD test: TPRs and FPRs on ImageNet-SCT and the top-$5\%$ DFM-filtered versions (w/ df).}
  \label{tab:ood}
  \centering
  \renewcommand{\arraystretch}{0.85}
  \begin{tabular}{p{1.9cm}p{0.9cm}c@{\hspace{16\tabcolsep}}c@{\hspace{8\tabcolsep}}c@{\hspace{8\tabcolsep}}c@{\hspace{8\tabcolsep}}c@{\hspace{8\tabcolsep}}c@{\hspace{8\tabcolsep}}c@{\hspace{8\tabcolsep}}c@{\hspace{8\tabcolsep}}c@{\hspace{8\tabcolsep}}c@{\hspace{8\tabcolsep}}c@{\hspace{8\tabcolsep}}c}
    \toprule
    \multicolumn{13}{c}{\bfseries ImageNet-SCT} \\
    \toprule
    \bfseries Model &  & & \bfseries military aircraft     & \bfseries car  & \bfseries lorikeet & \bfseries tabby cat & \bfseries holstein & \bfseries labrador retriever & \bfseries tree frog & \bfseries horse & \bfseries fishing vessel & \bfseries fire truck & \bfseries average \\
    \midrule
     \bfseries ResNet18 &&$TPR$  & 0.3286 & 0.4143 & 0.4429 & 0.2714 & 0.3286 & 0.4    & 0.4143 & 0.0286 & \textbf{0.4286} & 0.6143  &0.3672   \\
     && $FPR$  & 0.0794 & 0.0397 & 0.1952 & 0.0921 & 0.0746 & 0.0587 & 0.0429 & 0.019  & \textbf{0.0238} & 0.0778  & \\
     \cmidrule{2-14}
      & w/ df&$TPR$  &0      & 0      & 0.2143 & 0.1286 & 0.0429 & 0.0286 & 0.0571 & 0.1286 & \textbf{0.2143} & 0 &    \\
     && $FPR$ & 0.0016 & 0      & 0.054  & 0.0556 & 0.0683 & 0.0238 & 0.0063 & 0.0889 & \textbf{0.3397} & 0.0016    &    \\
    \midrule
    \bfseries ResNet18+AutoAug &&$TPR$  & 0.4    & 0.6571 & 0.5143 & 0.4    & 0.4857 & 0.4286 & 0.3286 & 0      & 0.4    & 0.6143 & 0.4229\\
     && $FPR$  & 0.0603 & 0.0667 & 0.1619 & 0.0937 & 0.1    & 0.0444 & 0.0302 & 0.0079 & 0.0143 & 0.0619 &   \\
     \cmidrule{2-14}
      & w/ df&$TPR$  & 0      & 0      & 0      & 0.0429 & 0.2143 & 0.0429 & 0.0143 & 0.0286 & 0.0429 & 0.0857 &     \\
     && $FPR$ & 0      & 0      & 0      & 0.0444 & 0.1016 & 0.0413 & 0      & 0.0079 & 0.0778 & 0.0127    &  \\
    \midrule
    \bfseries ResNet18+AugMix &&$TPR$  & 0.3571 & 0.7286 & 0.4143 & 0.2714 & 0.3857 & 0.4429 & 0.3571 & 0.0286 & \textbf{0.4143} & 0.5571 & 0.3957  \\
     && $FPR$  &0.0984 & 0.1159 & 0.1254 & 0.081  & 0.0889 & 0.054  & 0.0397 & 0.0111 & \textbf{0.0175} & 0.0397 & \\
     \cmidrule{2-14}
     & w/ df&$TPR$  & 0      & 0      & 0      & 0.1143 & 0.0429 & 0.2    & 0      & 0      & \textbf{0.5}    & 0.1429 &\\
      && $FPR$ &   0.0048 & 0      & 0.0095 & 0.0365 & 0.1    & 0.081  & 0      & 0.0111 & \textbf{0.2}    & 0.1016& \\
      \midrule
    \bfseries ResNet18+SIN &&$TPR$  & 0.3857& 0.6& 0.4286 & \textbf{0.4914}&0.6286 & 0.5714& 0.4571&0 &0.6429 &0.6857 & 0.48714\\
     && $FPR$  &0.0333 & 0.0444& 0.1016& \textbf{0.0476}& 0.1159& 0.0635& 0.0492&0.0222 &0.0127 &0.0794 & \\
     \cmidrule{2-14}
     & w/ df&$TPR$  &0.0429 &0 & 0.0714&\textbf{ 0.9286}& 0.0714& 0.1714 &0 &0 &0.0429 &0.0286&\\
      && $FPR$ &0.0349 &0.0016 &0.0222 &\textbf{0.7444 }& 0.0492 &0.1016 &0 &0.0159 &0.1127 & 0.0095& \\      
    \midrule
    \bfseries ResNet50 &&$TPR$  & 0.4286 & 0.4857 & 0.4143 & 0.2    & 0.3714 & 0.3    & 0.3    & 0.0571 & \textbf{0.4429} & 0.7429 &0.3743\\
        && $FPR$ &   0.1444 & 0.054  & 0.0952 & 0.0651 & 0.0984 & 0.0492 & 0.0365 & 0.027  & \textbf{0.0333} & 0.0921& \\
       \cmidrule{2-14}
        & w/ df&$TPR$  &  0.2429 & 0      & 0.0571 & 0.0429 & 0      & 0.2    & 0      & 0      &\textbf{ 0.4857} & 0.0429& \\
        && $FPR$ & 0.127  & 0      & 0.0032 & 0.0206 & 0      & 0.1444 & 0.0016 & 0.0159 &\textbf{ 0.3222} & 0.0111& \\
    \midrule
    \bfseries VGG16 &&$TPR$  & 0.5143 & 0.6571 & 0.4714 & 0.3    & 0.3571 & 0.3714 & 0.5143 & 0.0286 & \textbf{0.5286} & 0.5  &  0.4242\\
           && $FPR$ &  0.0841 & 0.0714 & 0.1238 & 0.073  & 0.0905 & 0.0492 & 0.0698 & 0.0143 & \textbf{0.0111} & 0.0524 &\\
            \cmidrule{2-14}
      & w/ df&$TPR$  & 0.0143 & 0      & 0.0286 & 0.2571 & 0.2143 & 0.1429 & 0.0143 & 0.0286 & \textbf{0.4571} & 0.0429& \\
         && $FPR$ &   0.0032 & 0      & 0.0032 & 0.2048 & 0.1079 & 0.0857 & 0      & 0.0333 &\textbf{ 0.4079} & 0.0571 &\\
    \bottomrule
  \end{tabular}
\end{table*}

 Learned frequency shortcuts might prevent NNs from learning meaningful semantics. We show an example of a person dressed in zebra-pattern clothes predicted as `zebra' with high confidence, and an image of a `horse' predicted as `zebra' with low confidence in~\cref{fig:zebracloth}. Mixing the images of `zebra cloth' and `horse' increases the confidence of being predicted as `zebra', indicating that the model mainly uses texture information and ignores almost any shape information of `zebra', potentially impairing generalization. As shown above, the class `zebra' is easily recognized early in the training, suggesting that learned frequency shortcuts impede the learning of other important semantics, e.g. the shape or other morphological features of the animal. 
 \textit{The learned frequency shortcuts are impacted significantly by the frequency characteristics of data. They can be texture-based or shape-based and  might hinder NNs from learning more meaningful semantics.} There might be cases where frequency shortcuts are not in the data and thus not learned. 

\paragraph{Model capacity vs. frequency shortcuts.}

The high TPR and FPR for ResNet50 in~\cref{tab:noFrqs} indicate that it is subject to frequency shortcuts for the classification of classes `airliner' and `container ship'. 
Compared to ResNet18 frequency shortcut for class `zebra', ResNet50 has lower TPR and FPR, indicating less specific dominant frequencies for classifying `zebra'. This demonstrates mitigation of learning a frequency shortcut, although learning another shortcut for class `airliner'.  Additionally, VGG16 learns a frequency shortcut for class `container ship' (TPR=0.7 and FPR=0.42). We show in the following paragraph that frequency shortcuts affect transformers as well, indicating that shortcuts impact networks across different model capacities and architectures. Thus, larger models cannot necessarily avoid it. 
This commonality shows that frequency shortcut learning is data-driven, which needs to be considered more explicitly to learn generalizable models.

\paragraph{Transferability of frequency shortcuts.}
We trained ViT-B on ImageNet-10 and tested it on images processed with the DFMs we had computed for ResNet18+SIN. This tests the dependency of ViT predictions on small sets of frequency, and the transferability of shortcuts between models or architectures.  We present the results in~\cref{tab:vitb} and observe shortcuts  for the classes `siamese cat' (TPR=0.82, FPR=0.22) and `container ship' (TPR=0.92, FPR=0.25).  
Though having a large model capacity, ViT-B is also subject to frequency shortcuts (shape or texture) to classify the samples of certain classes, in line with the observation in~\cite{9783031}. Moreover, the frequency shortcuts learned by ResNet18+SIN can be exploited by ViT-B, further indicating that frequency shortcuts are data-driven and can be transferred between models.

\paragraph{Data augmentation vs. frequency shortcuts.}
\label{para:dataaug}

As common techniques to improve generalization performance, we investigate the effect of data augmentation in mitigating frequency shortcut learning. We train ResNet18 with these techniques and report the results in~\cref{tab:noFrqs}. 
 AugMix worsens the learned frequency shortcut for `container ship', but mitigates a frequency shortcut for `zebra'. AutoAugment partially avoids the frequency shortcuts for both `zebra' and `container ship'. SIN causes a frequency shortcut for `siamese cat'. To summarize, appropriate data augmentation may partially reduce frequency shortcut learning, but NNs  still tend to find shortcut solutions based on the characteristics of the augmented data.
 

\section{Frequency shortcuts and OOD tests} 
\label{sec:oodtest}
\paragraph{Design of OOD test: ImageNet-SCT.}
To assess how frequency shortcuts affect OOD generalization, we construct a new test set based on previous analysis results, ImageNet-SCT (ShortCut Tests). It consists of 10 classes, each containing 70 images with seven different image styles, including \emph{art, cartoon, deviantart, painting, sculpture, sketch, toy}. This dataset expands the coverage of ImageNet-R~\cite{HendrycksICCV21} in terms of image variations. 
The classes in ImageNet-SCT are related, to some extent, to those in ImageNet-10. For instance, `zebra' in ImageNet-10 corresponds to `horse' in ImageNet-SCT, allowing us to test the effect of an absent texture-based shortcut feature, as horse images contain animals with a very similar shape to zebras, but with no texture. Similarly, 'siamese cat' in ImageNet-10 corresponds to 'tabby cat' in ImageNet-SCT, to test the effect of a present shape-based shortcut feature.  Furthermore, `container ship' in ImageNet-10 maps to `fishing vessel' in ImageNet-SCT, which contains images with similar textures and somehow different shapes (fishing vessels are much smaller boats), enabling us to evaluate the effect of a present texture-based shortcut. Examples of ImageNet-SCT images are provided in the supplementary material.

\paragraph{Frequency shortcuts can impair generalization and create the illusion of improved performance.}
We test the NNs on ImageNet-SCT and its DFM-filtered versions with the top-$5\%$ dominant frequencies. From the results on the original ImageNet-SCT, we observe a considerable average drop of TPR for all models (see~\cref{tab:ood}). 
Larger model capacity and data augmentations may not always effectively address frequency shortcuts in certain classes, as observed for `siamese cat', `zebra', and `container ship' in ImageNet-10 (corresponding to `tabby cat', `horse', and `fishing vessel' in ImageNet-SCT). For example, models relying on texture-based shortcut features for `zebra' in ImageNet-10 fail to capture shape characteristics and perform poorly on similar-shaped animals like `horse' in ImageNet-SCT (see~\cref{fig:zebracloth}). While data augmentations can partially mitigate this effect in ID tests, OOD results for `horse' still indicate the presence of learned frequency shortcuts. Conversely, `tabby cat' and `fishing vessel', which are designed to have similar shape or texture characteristics to their corresponding class in ImageNet-10,  exhibit above-average OOD results (higher TPR than average accuracy). Thus, the present shape-based and texture-based shortcut features in the OOD test set are used for classification, giving a false sense of generalization. `Fire truck' in ImageNet-SCT is a good example of generalization, as no shortcuts were identified, allowing models to learn more global and semantic information. 
Frequency shortcuts can impair generalization and their impact can transfer across datasets, resulting in a misleading impression of generalization with the inclusion of shortcut features in a new test set. Larger models and data augmentation cannot fully counteract these effects, we thus highlight the need to explore novel data augmentation strategies that explicitly target shortcut mitigation, e.g. leveraging DFMs to induce models to exploit more frequencies rather than shortcut frequencies~\cite{wang2023dfmx} and avoid learning behaviors that may impair the generalizability of NNs.

\section{Conclusions}
\label{conclusion}
We conducted an empirical study to investigate what NNs learn in image classification, by analyzing the learning dynamics of NNs from a frequency shortcut perspective. We found from a synthetic example that \textbf{NNs learn frequency shortcuts during training to simplify classification tasks, driven by frequency characteristics of data and simplicity-bias}. To address this on natural images, we proposed a metric to measure class-wise frequency characteristics and a method to identify frequency shortcuts. We evaluated the influence of shortcuts on OOD generalization and found that \textbf{frequency shortcuts can be transferred to another dataset, in some cases, giving an illusion of improved generalization}. Furthermore, we observed that larger model capacity and data augmentation techniques do not necessarily mitigate frequency shortcut learning.
Our study expands previous works on the learning dynamics of NNs for regression tasks, broadens the understanding of frequency shortcuts (which can be either texture-based or shape-based), and provides a more systematic analysis of OOD generalization. We foresee that enhancing the identification of frequency shortcuts and applying proper training schemes that avoid frequency shortcut learning may hold promise in improving generalization.

\subsection*{Acknowledgements}
\noindent This work was supported by the SEARCH project (\url{https://sites.google.com/view/search-utwente}), UT Theme Call 2020, Faculty of Electrical Engineering, Mathematics and Computer Science, University of Twente.

{\small
\bibliographystyle{ieee_fullname}
\bibliography{egbib}
}

\clearpage
 \appendix
 \input{supplementary}

\end{document}

%% file: imgs/zebracloth_new.tex
\begingroup%
  \makeatletter%
  \providecommand\color[2][]{%
    \errmessage{(Inkscape) Color is used for the text in Inkscape, but the package 'color.sty' is not loaded}%
    \renewcommand\color[2][]{}%
  }%
  \providecommand\transparent[1]{%
    \errmessage{(Inkscape) Transparency is used (non-zero) for the text in Inkscape, but the package 'transparent.sty' is not loaded}%
    \renewcommand\transparent[1]{}%
  }%
  \providecommand\rotatebox[2]{#2}%
  \newcommand*\fsize{\dimexpr\f@size pt\relax}%
  \newcommand*\lineheight[1]{\fontsize{\fsize}{#1\fsize}\selectfont}%
  \ifx\svgwidth\undefined%
    \setlength{\unitlength}{245bp}%
    \ifx\svgscale\undefined%
      \relax%
    \else%
      \setlength{\unitlength}{\unitlength * \real{\svgscale}}%
    \fi%
  \else%
    \setlength{\unitlength}{\svgwidth}%
  \fi%
  \global\let\svgwidth\undefined%
  \global\let\svgscale\undefined%
  \makeatother%
  \begin{picture}(1,0.35376204)%
    \lineheight{1}%
    \setlength\tabcolsep{0pt}%
    \put(0,0){\includegraphics[width=\unitlength,page=1]{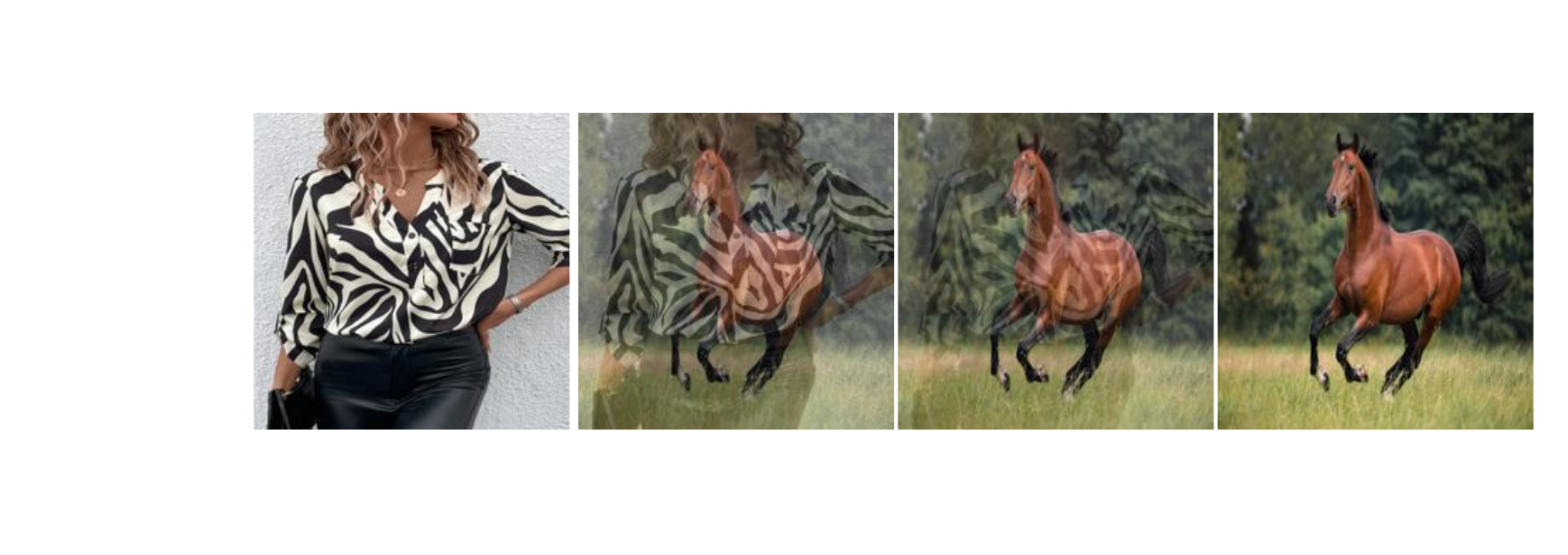}}%
    \put(0.15745503,0.30026853){\color[rgb]{0,0,0}\makebox(0,0)[lt]{\lineheight{1.25}\smash{\scriptsize\begin{tabular}[t]{l}100\% zebra cloth\end{tabular}}}}%
    \put(0.80623599,0.30026853){\color[rgb]{0,0,0}\makebox(0,0)[lt]{\lineheight{1.25}\smash{\scriptsize\begin{tabular}[t]{l}100\% horse\end{tabular}}}}%
    \put(0.37936829,0.340643071){\color[rgb]{0,0,0}\makebox(0,0)[lt]{\lineheight{1.25}\smash{\scriptsize\begin{tabular}[t]{l}{\color{RubineRed}40\%} zebra cloth\end{tabular}}}}%
    \put(0.581441995,0.340643071){\color[rgb]{0,0,0}\makebox(0,0)[lt]{\lineheight{1.25}\smash{\scriptsize\begin{tabular}[t]{l}{\color{RubineRed}20\%} zebra cloth\end{tabular}}}}%
    \put(0.026,0.04){\color[rgb]{0,0,0}\makebox(0,0)[lt]{\lineheight{1.25}\smash{\scriptsize\begin{tabular}[t]{l}Zebra\end{tabular}}}}%
    \put(0.2303865,0.04){\color[rgb]{0,0,0}\makebox(0,0)[lt]{\lineheight{1.25}\smash{\scriptsize\begin{tabular}[t]{l}99.84\%\end{tabular}}}}%
    \put(0.43734712,0.04){\color[rgb]{0,0,0}\makebox(0,0)[lt]{\lineheight{1.25}\smash{\scriptsize\begin{tabular}[t]{l}97.26\%\end{tabular}}}}%
    \put(0.64126261,0.04){\color[rgb]{0,0,0}\makebox(0,0)[lt]{\lineheight{1.25}\smash{\scriptsize\begin{tabular}[t]{l}21.23\%\end{tabular}}}}%
    \put(0.85066892,0.04){\color[rgb]{0,0,0}\makebox(0,0)[lt]{\lineheight{1.25}\smash{\scriptsize\begin{tabular}[t]{l}2.11\%\end{tabular}}}}%
    \put(0.05,0.001){\color[rgb]{0,0,0}\makebox(0,0)[lt]{\lineheight{1.25}\smash{\scriptsize\begin{tabular}[t]{l}Ox\end{tabular}}}}%
    \put(0.24952044,0.001){\color[rgb]{0,0,0}\makebox(0,0)[lt]{\lineheight{1.25}\smash{\scriptsize\begin{tabular}[t]{l}0\%\end{tabular}}}}%
    \put(0.44281396,0.001){\color[rgb]{0,0,0}\makebox(0,0)[lt]{\lineheight{1.25}\smash{\scriptsize\begin{tabular}[t]{l}0.74\%\end{tabular}}}}%
    \put(0.64126261,0.001){\color[rgb]{0,0,0}\makebox(0,0)[lt]{\lineheight{1.25}\smash{\scriptsize\begin{tabular}[t]{l}23.87\%\end{tabular}}}}%
    \put(0.84520208,0.001){\color[rgb]{0,0,0}\makebox(0,0)[lt]{\lineheight{1.25}\smash{\scriptsize\begin{tabular}[t]{l}93.73\%\end{tabular}}}}%
    \put(0.0,0.17237327){\color[rgb]{0,0,0}\makebox(0,0)[lt]{\lineheight{1.25}\smash{\scriptsize\begin{tabular}[t]{l}Predicted  \end{tabular}}}}%
    \put(0.05,0.14237327){\color[rgb]{0,0,0}\makebox(0,0)[lt]{\lineheight{1.25}\smash{\scriptsize\begin{tabular}[t]{l} as \end{tabular}}}}%
    \put(0.4064828,0.29666138){\color[rgb]{0,0,0}\makebox(0,0)[lt]{\lineheight{1.25}\smash{\scriptsize\begin{tabular}[t]{l}60\% horse\end{tabular}}}}%
    \put(0.61838533,0.29666138){\color[rgb]{0,0,0}\makebox(0,0)[lt]{\lineheight{1.25}\smash{\scriptsize\begin{tabular}[t]{l}80\% horse\end{tabular}}}}%
  \end{picture}%
\endgroup%

%% file: supplementary.tex
\section{Datasets}
\label{sup:datasets}
\subsection{Synthetic datasets}
\begin{figure}[h]
    \centering
    \begin{subfigure}[b]{0.5\textwidth}
         \centering
         \includegraphics[width=\textwidth]{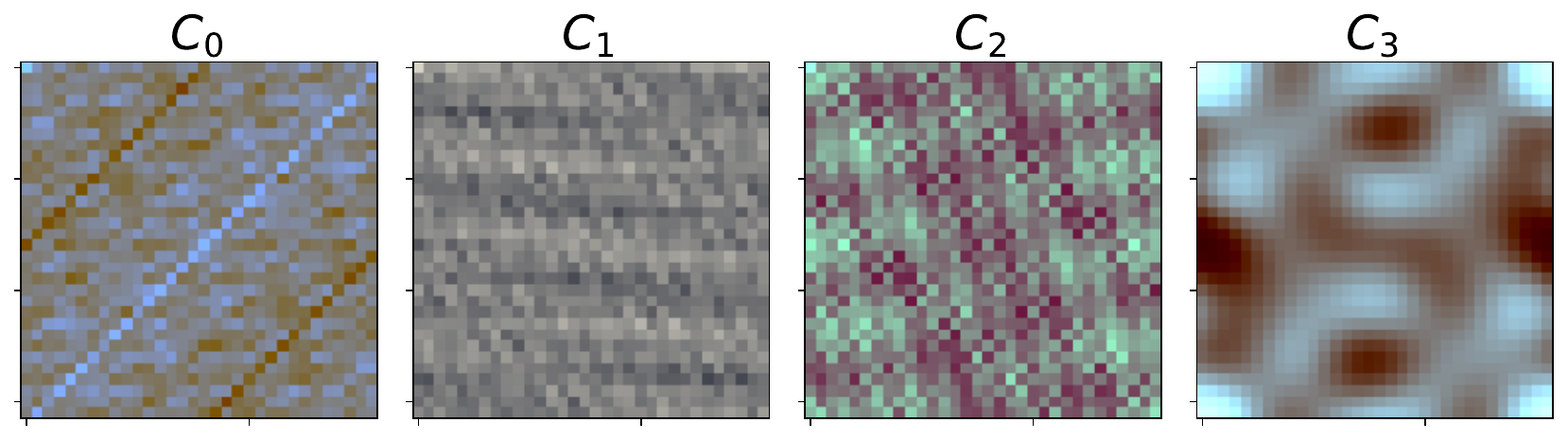}
         \caption{$Syn_{B_1}$}
         \label{fig:syn1}
     \end{subfigure}
     \begin{subfigure}[b]{0.5\textwidth}
         \centering
         \includegraphics[width=\textwidth]{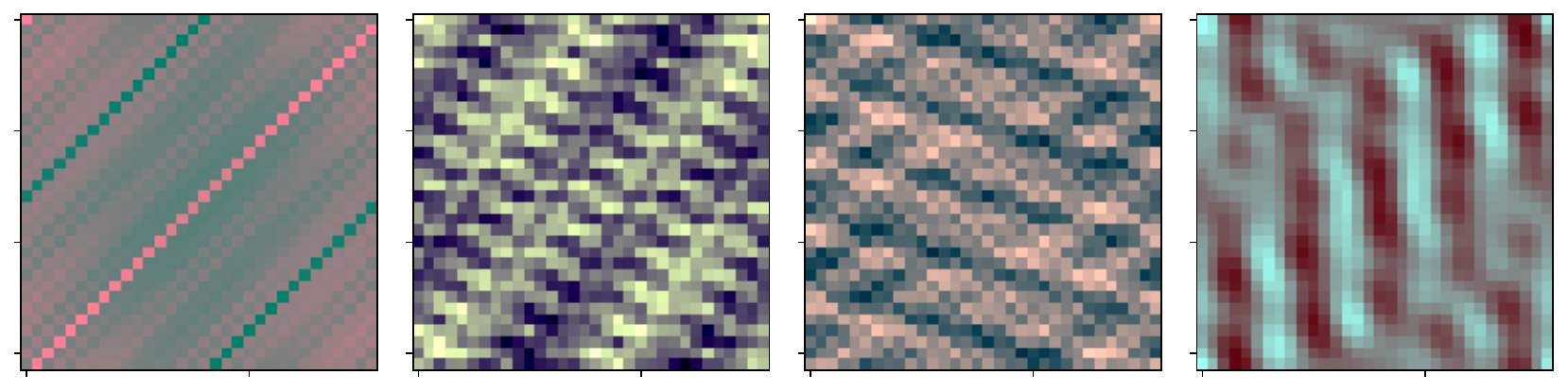}
         \caption{$Syn_{B_2}$}
         \label{fig:syn2}
     \end{subfigure}
     \begin{subfigure}[b]{0.5\textwidth}
         \centering
         \includegraphics[width=\textwidth]{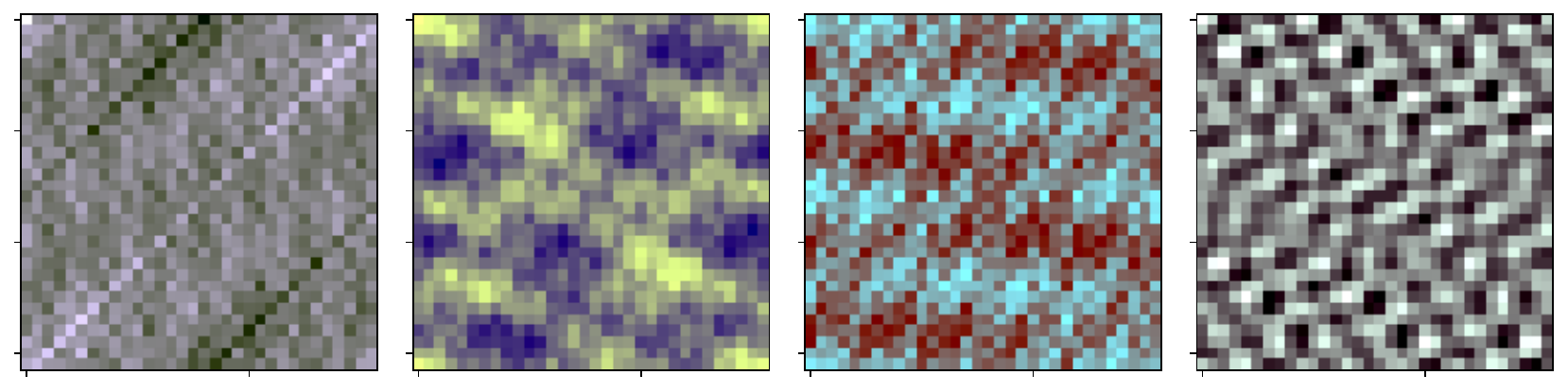}
         \caption{$Syn_{B_3}$}
         \label{fig:syn3}
     \end{subfigure}
     \begin{subfigure}[b]{0.5\textwidth}
         \centering
         \includegraphics[width=\textwidth]{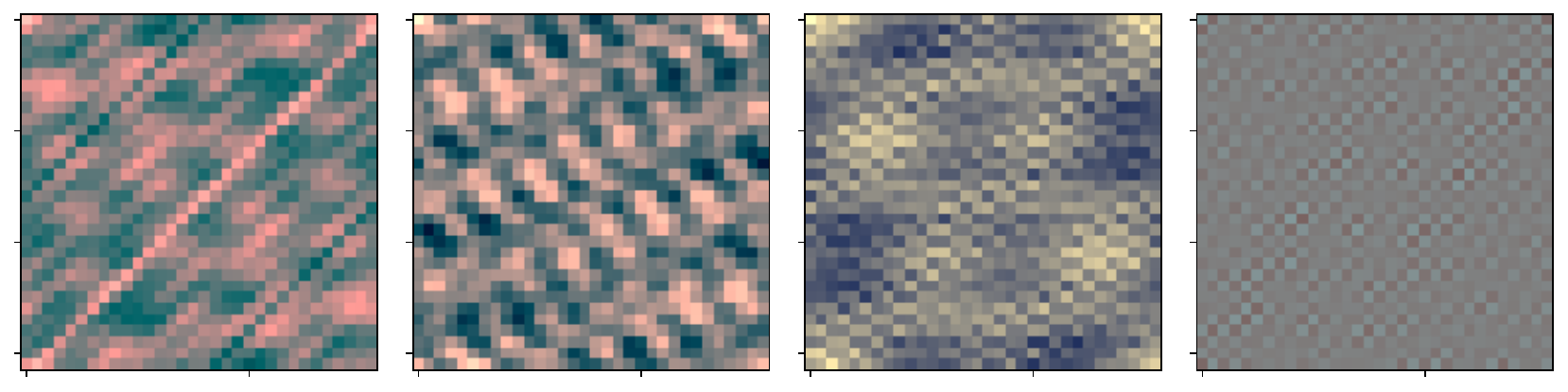}
         \caption{$Syn_{B_4}$}
         \label{fig:syn4}
     \end{subfigure}
    \caption{\small{Samples of synthetic datasets. Class $C_3$ has a frequency bias to a specific band, which is $B_1$ for $Syn_{B_1}$, $B_2$ for $Syn_{B_2}$, $B_3$ for $Syn_{B_3}$, and $B_4$ for $Syn_{B_4}$.
    Due to frequency bias, images of class $C_3$ can be easily distinguished from other classes. }}
    \label{fig:synsamples}
\end{figure}

The frequency bands of each class in synthetic datasets are shown in~\cref{tab:detailsbandsyn}. For each dataset, class $C_3$ has a bias to a specific band, and classes $C_0$ and $C_1$ are designed to contain frequencies from the other three bands. Class $C_2$ contains frequencies across the whole spectrum. 
Example images from the four synthetic datasets that we created are shown in~\cref{fig:synsamples}. We designed the classes so that they have specific frequency characteristics. We induced different levels of class-wise difficulty when the NNs are trained to distinguish their samples.  Across the four datasets, the images of class $C_3$ are easily distinguishable from those of the other three classes, as observed visually. This is because class $C_3$  has a frequency bias to a specific band, e.g.  low-frequency bias in the Syn$_{B_1}$ dataset and high-frequency bias in the Syn$_{B_4}$ dataset.
The images of classes $C_0$, $C_1$, and $C_2$ are visually similar across the four synthetic datasets. Despite the visual similarity, the images of class $C_0$ have \emph{special patterns} consisting of a fixed set of frequencies across the spectrum. The \emph{special patterns} are the designed characteristics making the images of class $C_0$ easily distinguishable from classes $C_1$ and $C_2$. Note that, the \emph{special patterns} consist of eight frequencies that can be evenly filtered based on the band-stop filters we use during testing. This is to analyze how the NNs utilize frequency information from the \emph{special patterns} fairly. 
The difference between classes $C_1$ and $C_2$ is the number of frequency bands sampled for the data generation.  Class $C_1$ has one less sampling band than those of class $C_2$. However, for the images of classes $C_1$ and $C_2$, it is hard for human observers to identify their difference visually while NNs can, according to their classification results. On the other hand, classes $C_0$ and $C_3$ are easier for human observers to be visually distinguished. 

\begin{table*}
  \caption{Frequency bands of each synthetic dataset. }
  \label{tab:detailsbandsyn}
  \centering
  \renewcommand{\arraystretch}{1.1}
  \footnotesize
  \begin{tabular}{ccccc}
    \toprule
    & \multicolumn{4}{c}{\bfseries Synthetic datasets} \\ \midrule
    \bfseries class &    $Syn_{B_1}$ &$Syn_{B_2}$ &$Syn_{B_3}$ &$Syn_{B_4}$ \\
    \midrule
     \bfseries $C_0$ & $B_2,B_3,B_4$  &$B_1,B_3,B_4$&$B_1,B_2,B_4$&$B_1,B_2,B_3$ \\
     \bfseries $C_1$ & $B_2,B_3,B_4$ &$B_1,B_3,B_4$&$B_1,B_2,B_4$&$B_1,B_2,B_3$ \\
      \bfseries $C_2$ & $B_1,B_2,B_3,B_4$  &$B_1,B_2,B_3,B_4$&$B_1,B_2,B_3,B_4$&$B_1,B_2,B_3,B_4$ \\
      \bfseries $C_3$ & $B_1$ &$B_2$&$B_3$&$B_4$ \\
    \bottomrule
  \end{tabular}
\end{table*}

\subsection{OOD test data: ImageNet-SCT}
\begin{figure*}
     \centering
     \input{imgs/sc_test_new}
     \caption{Example images from the ImageNet-SCT dataset. Images are organized in 10 classes, with images of seven different renditions: (in order of the columns) art, cartoon, deviantart, painting, sculpture, sketch, and toy.}
     \label{fig:samplesctest}
\end{figure*}
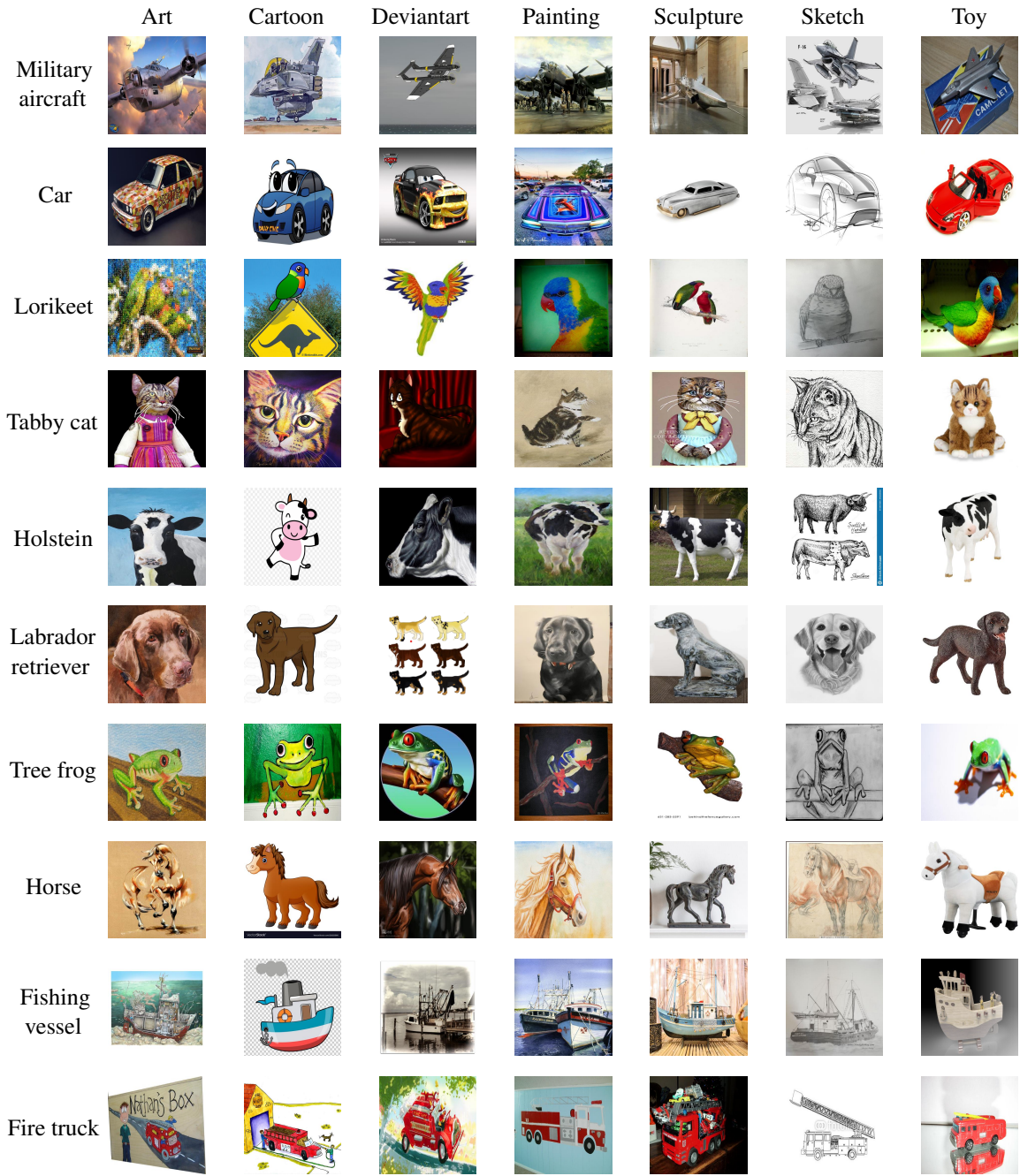
ImageNet-SCT is specifically designed to validate the influence of frequency shortcuts on an unseen dataset, for models trained on ImageNet10. The analysis demonstrates that NNs might learn frequency shortcuts for easier classification, which correspond to texture-based or shape-based patterns. The classification dependency on the patterns shows that NNs might ignore other useful semantics. Therefore, to validate how this learning behavior affects OOD generalization, we construct a new dataset, containing 10 classes similar to those of ImageNet-10~\cite{huang2021unlearnable} but with different shape/texture characteristics. ImageNet-trained NNs are found to have a texture bias~\cite{geirhos2018imagenettrained}. Thus, the main criterion applied for the composition of the dataset is to have classes with similar shape characteristics to ImageNet-10, instead of texture characteristics, except for classes  `military aircraft', `car', and `fishing vessel' which have similar texture characteristics to the corresponding classes in ImageNet-10. This helps to evaluate the influence of learned frequency shortcuts on an OOD test from two perspectives, namely when the shortcut features are present or absent.   Each class contains 7 renditions of images (i.e. art, cartoon, deviantart, painting, sculpture, sketch and toy), which is inspired by the design idea of ImageNet-R~\cite{HendrycksICCV21}. Example images of ImageNet-SCT are shown in~\cref{fig:samplesctest}. Each row shows the images of the seven renditions of one class.
\begin{figure*}[!t]
     \centering
     \includegraphics[width = 0.75\linewidth]{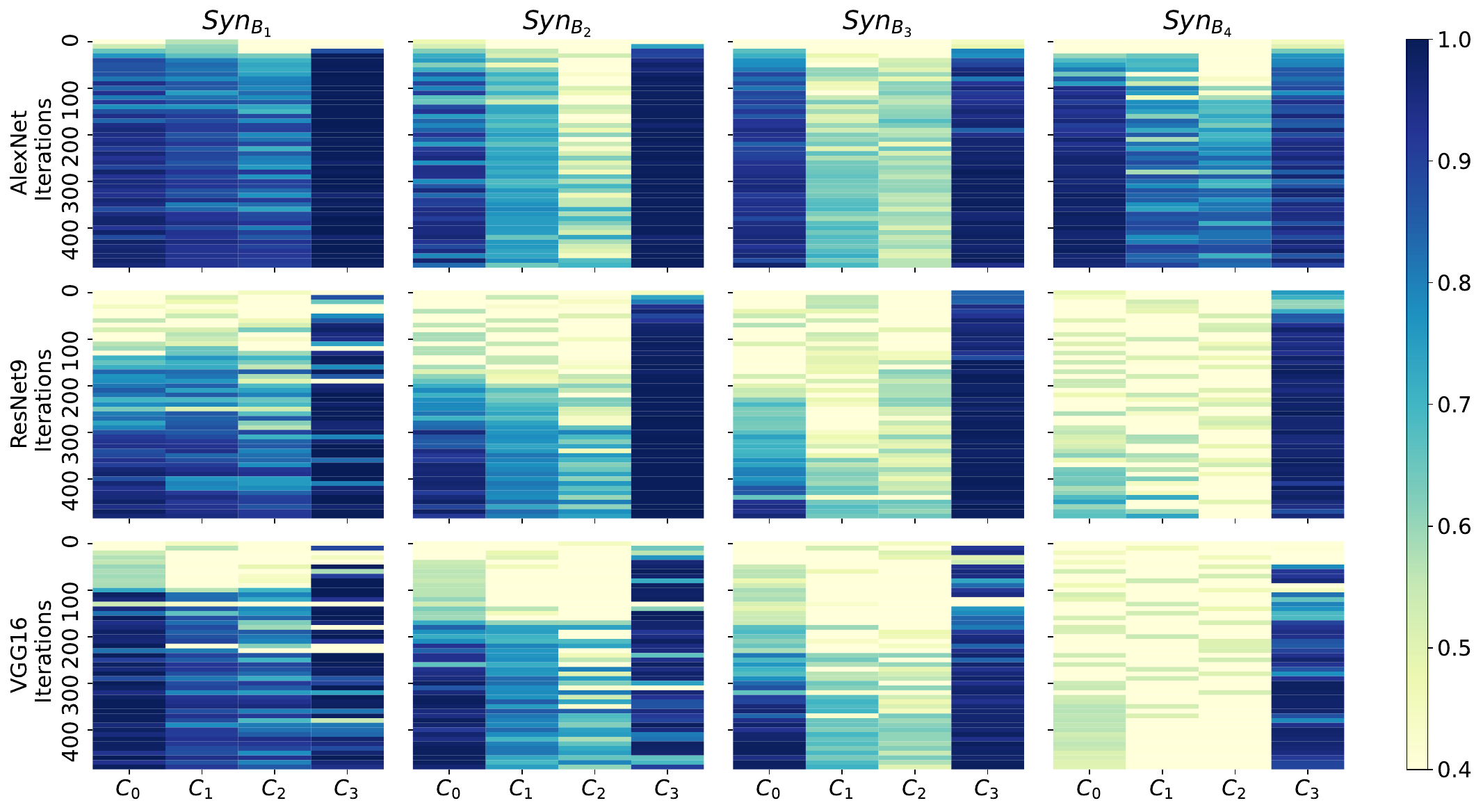}
     \caption{F$_1$-scores of the first 500 iterations of AlexNet, ResNet9, and VGG16 trained on the four synthetic datasets respectively. }
     \label{fig:F1_Syn_vggalex}
\end{figure*}
\section{Training setup}
\paragraph{Synthetic datasets.}
We train AlexNet~\cite{NIPS2012_c399862d}, ResNet(s)~\cite{151203385} and VGG-16~\cite{14091556} models  for 100 epochs on the four synthetic datasets. The initial learning rate is 0.01, reduced by a factor of 10 if the validation loss does not decrease for 10 epochs. We use SGD optimizer with momentum $0.9$ and weight decay $10^{-4}$, and  batch size $128$.

\paragraph{ImageNet-10 dataset.}
Models with ResNet(s)~\cite{151203385} and VGG-16~\cite{14091556} architectures are trained for 200 epochs on the ImageNet-10 dataset. The initial learning rate is 0.01 and is reduced by a factor of 10 if the validation loss does not decrease for 10 epochs. We use SGD optimizer with momentum $0.9$ and weight decay $10^{-4}$, and batch size $16$.

\section{Extra results}
\subsection{Synthetic datasets.}
\paragraph{F$_1$-scores}
\cref{fig:F1_Syn_vggalex} shows the $F_1$-score computed on the test sets of the four synthetic datasets during the first 500 iterations of the training of AlexNet, ResNet9 and VGG16. As generally observable, all model architectures achieve higher $F_1$-scores for class $C_3$ than for the other classes. This indicates that class $C_3$  is recognized immediately and easily by the NNs during training. This is consistent with the results of ResNet18 and shows the existence of shortcut learning, which prioritizes the recognition of easily distinguishable frequency patterns.

\begin{figure}[!t]
     \centering
     \includegraphics[width = \linewidth]{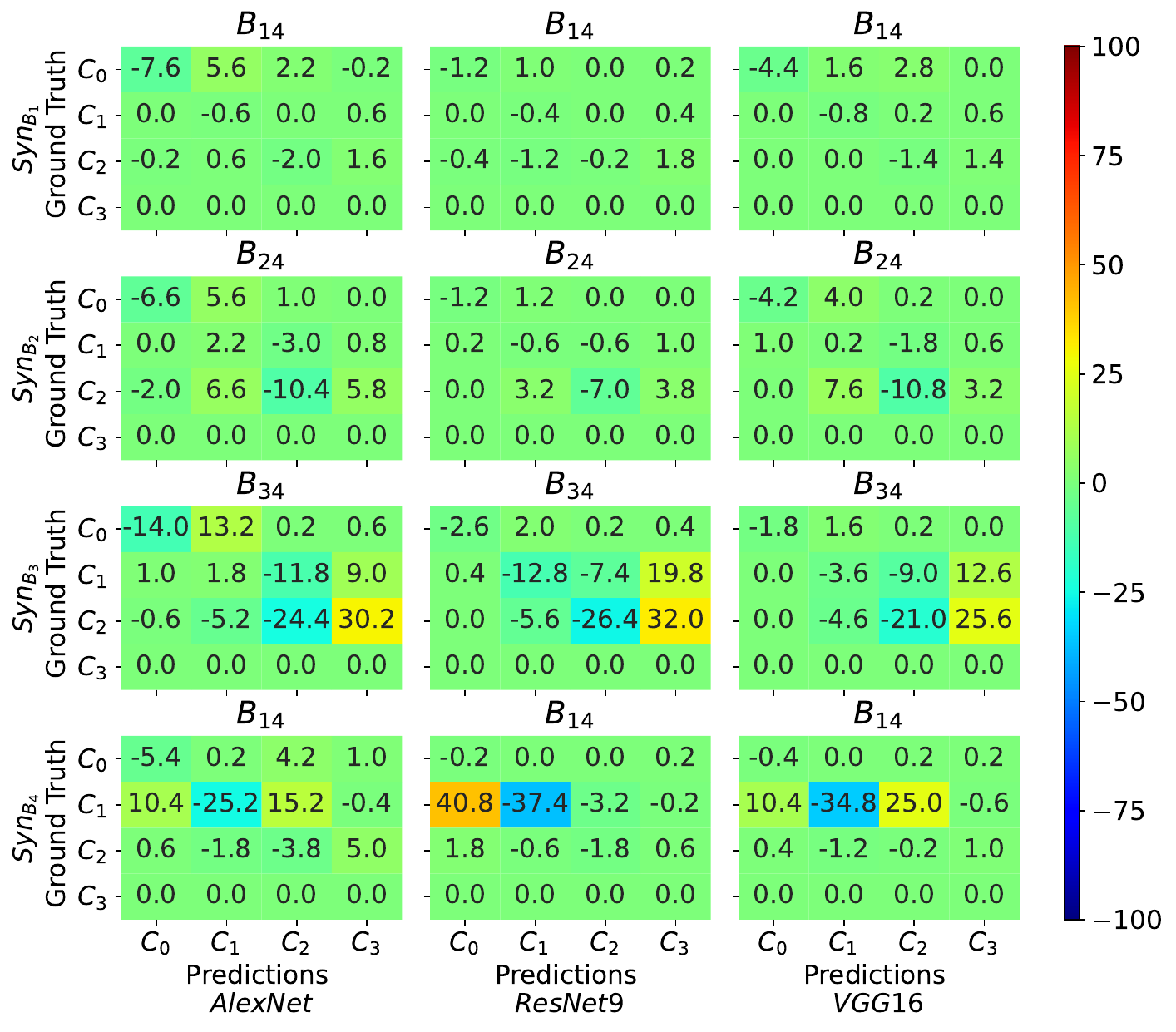}
     \caption{Relative confusion matrices of AlexNet, ResNet9, and VGG16 trained on the synthetic datasets }
     \label{fig:conf_Syn_vggalex}
\end{figure}

\paragraph{Relative confusion matrices.}
\cref{fig:conf_Syn_vggalex} shows the relative confusion matrices of AlexNet (first column), ResNet9 (second column) and VGG16 (third column) trained on the four synthetic datasets. The models are tested on the different band-stop test sets, obtained by suppressing in turn the frequencies in two out of the four sub-bands considered for the data generation. Because of the class-wise frequency characteristics of the synthetic datasets, these tests are meant to inspect the frequency utilization of different NN models, i.e. what frequencies are needed for classification. 
The performance results of the models are mostly stable when they are tested on test sets retaining only two frequency bands (see the values of $\Delta^{C_i,C_i}$ where $C_i \in \{C_0,C_1,C_2,C_3\}$), showing that they do not need complete frequency information for classification. For instance, class $C_0$ has a \emph{special pattern} consisting of frequencies across the whole spectrum, and the corresponding $\Delta^{C_0,C_0}$ is mostly close to zero.
Models may find shortcut solutions in the Fourier domain for classification and this behavior is common across different architectures.


\begin{figure}[!t]

    \centering
    \includegraphics[width = \linewidth]{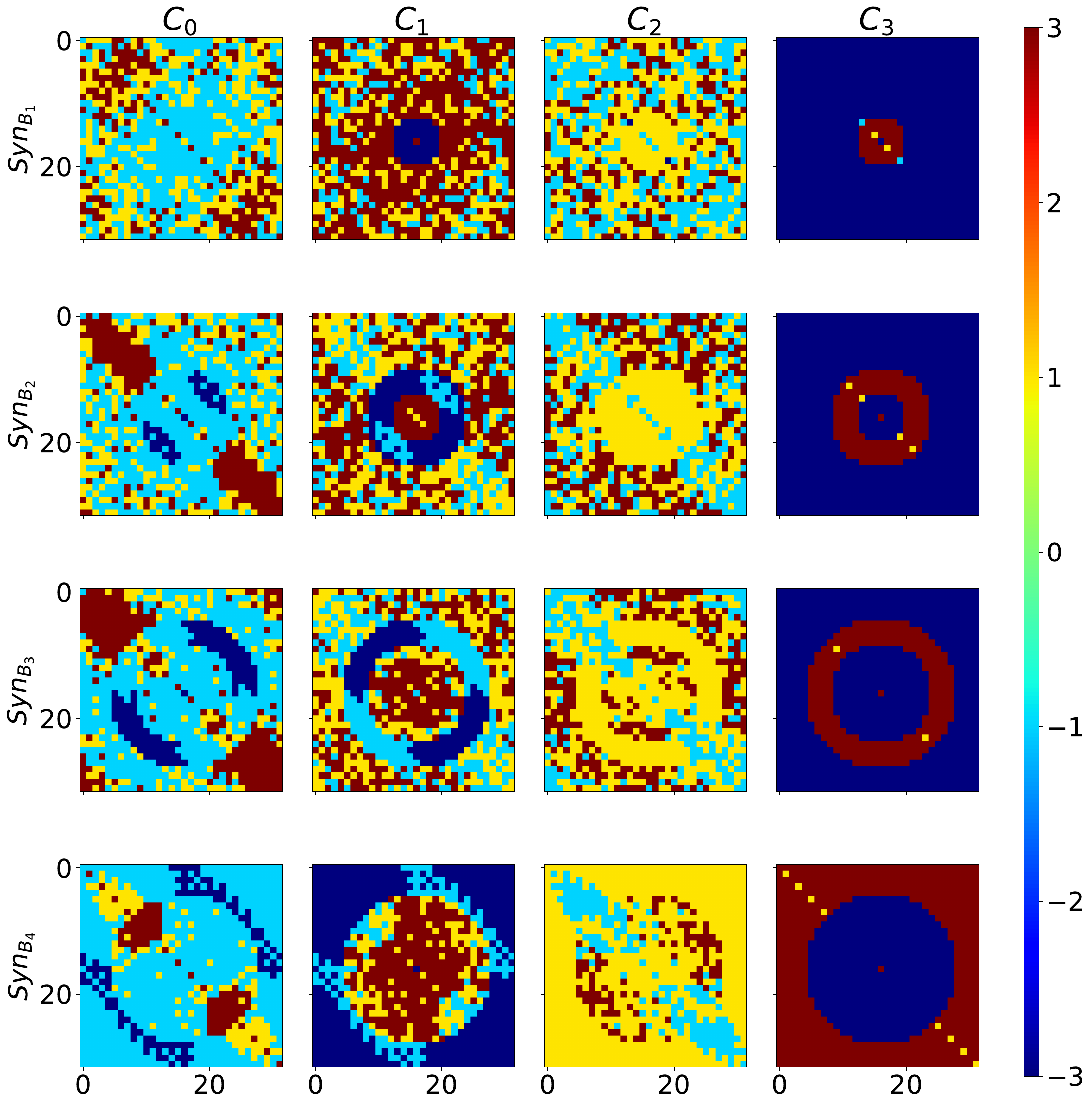}
    \caption{ADCS of the classes in synthetic datasets.}
    \label{fig:adcssynthetic}
\end{figure}

\paragraph{ADCS.}
The ADCS of classes in the four synthetic datasets are shown in~\cref{fig:adcssynthetic}. Across the four datasets, class $C_3$ has a significant bias on a specific band, from low to high. The yellow dots in $ADCS^{C_3}$ (belong to some frequencies in the frequency set of the \textit{special\ pattern}) indicate that the corresponding frequencies have slightly more energy  than other classes, which is caused by the removal of the specific frequencies (non-ideal filtering). Class $C_0$ has more energy around the specific frequency sets than other classes, this is also due to the non-ideal filtering. In general, the ADCS shows that the classes in a synthetic dataset $Syn_b$ have distinguishable frequency characteristics. These might be used as shortcuts. The class with the most distinctive frequency characteristics, i.e. class $C_3$ is learned first by NNs in the training phase (see~\cref{fig:F1_Syn_vggalex}), indicating that the models have a tendency to identify that distinctive frequency characteristic as an easy solution for the classification problem. ADCS can be used to analyze the class-wise frequency characteristics in a dataset, rather than being used directly to predict which class might be learned first. Further investigation on frequency characteristics and learning dynamics  is needed to establish if certain frequency characteristics induce a shortcut or not.

\begin{figure}[!t]
    \centering
\subfloat[airliner]{\includegraphics[height=2.2cm]{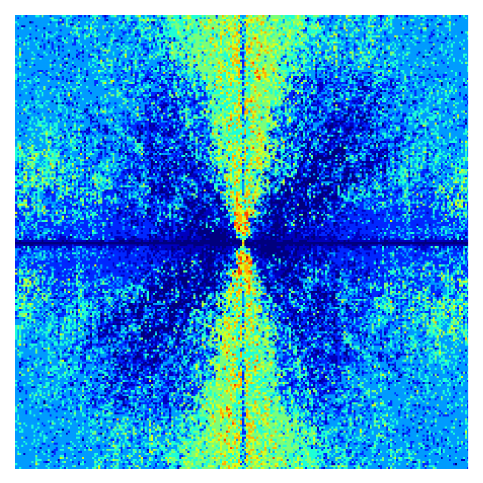}}
\subfloat[wagon]{\includegraphics[height=2.2cm]{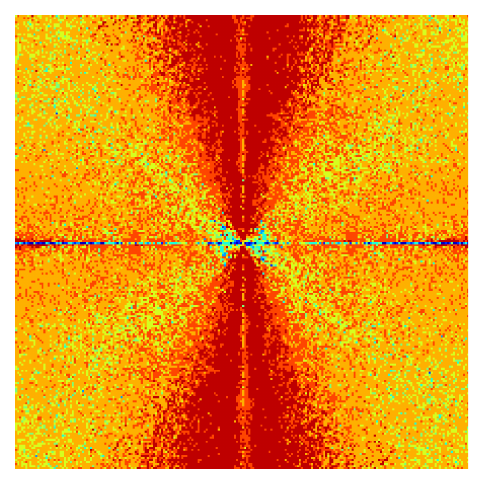}}
\subfloat[siamese cat]{\includegraphics[height=2.2cm]{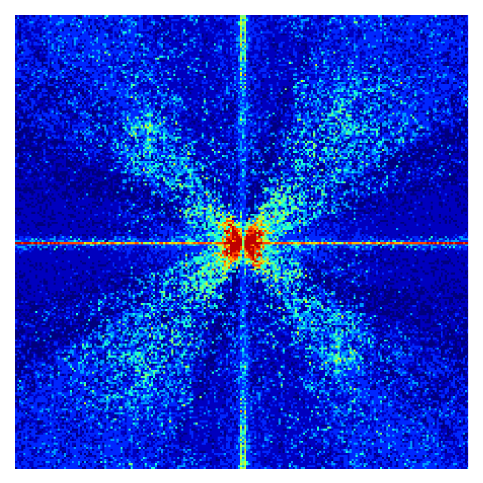}}
\subfloat[ox]{\includegraphics[height=2.2cm]{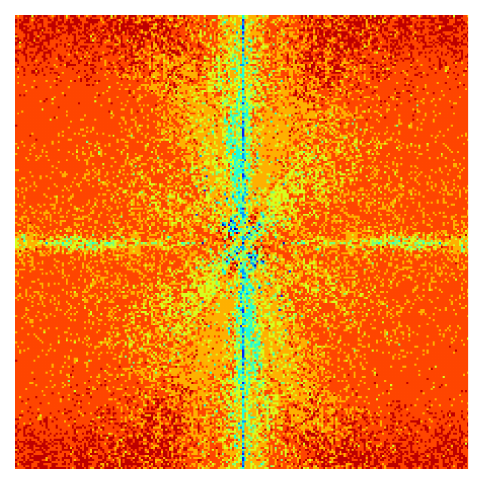}}\\
\subfloat[golden retriever]{\includegraphics[height=2.2cm]{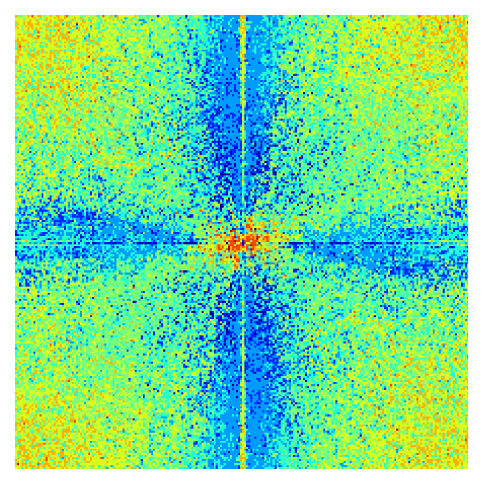}}
\subfloat[tailed frog]{\includegraphics[height=2.2cm]{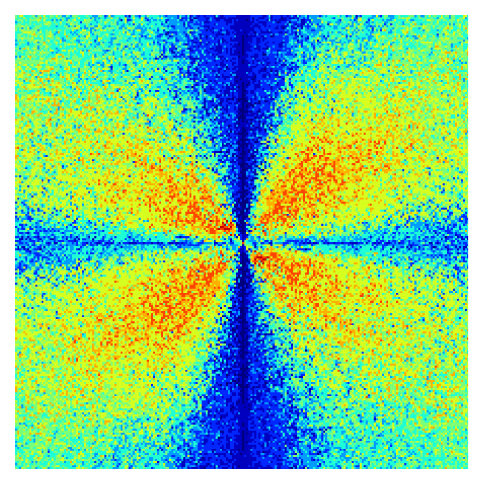}}
\subfloat[container ship]{\includegraphics[height=2.2cm]{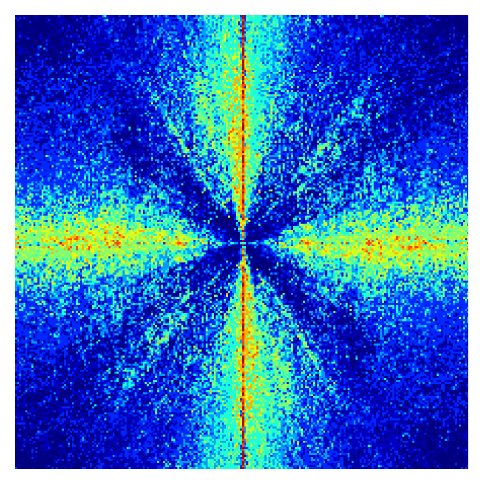}}
\subfloat[trailer truck]{\includegraphics[height=2.2cm]{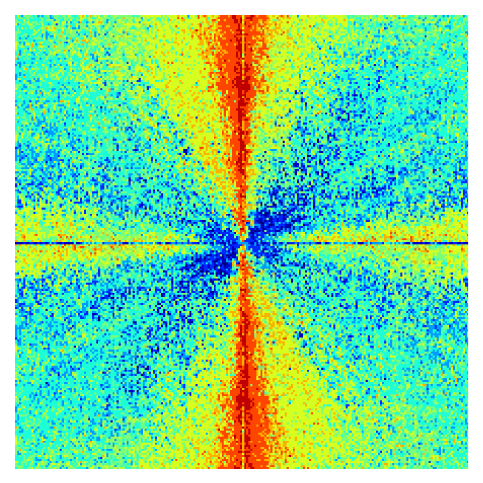}}
\vspace{-2mm}
    \caption{ADCS of other classes in ImageNet-10.}
    \label{fig:adcsimagenet}
\end{figure}

\begin{figure}[!t]
    \centering
     \begin{subfigure}[b]{0.5\textwidth}
        \centering
        \includegraphics[width = \textwidth]{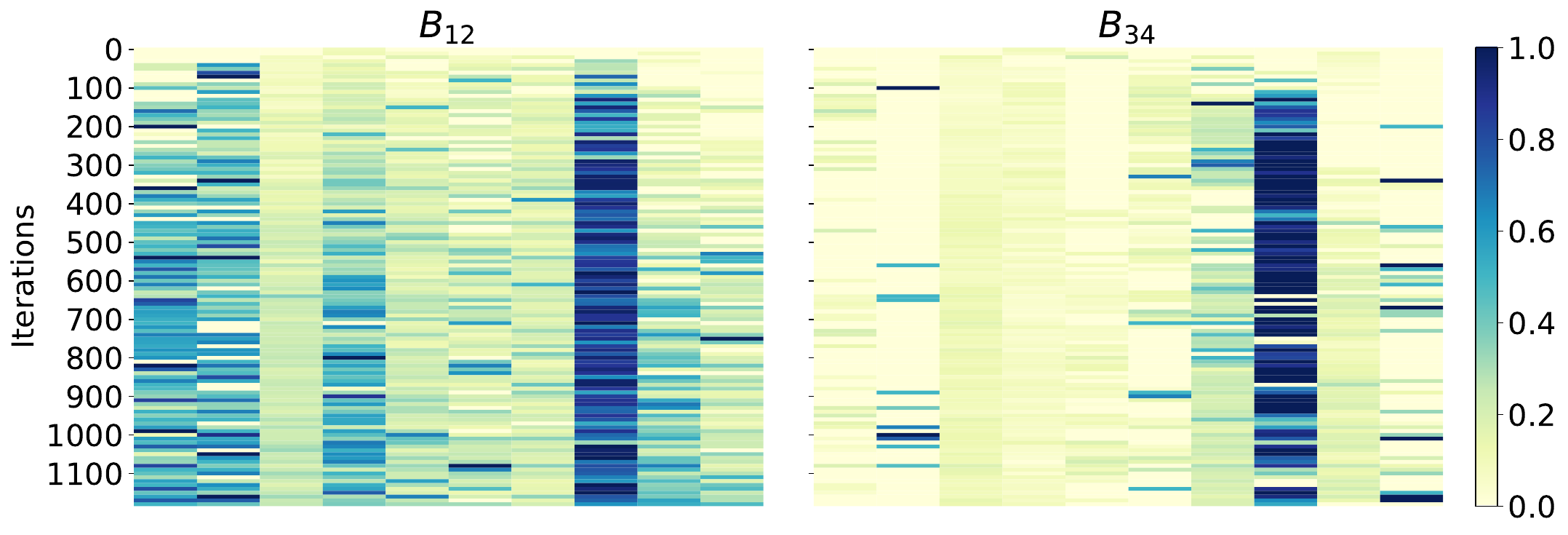}
        \caption{Precision of ResNet50.}
        \label{fig:learimagenet10Precision_resnet50}
        \end{subfigure}
    \begin{subfigure}[b]{0.5\textwidth}
        \centering
        \includegraphics[width = \textwidth]{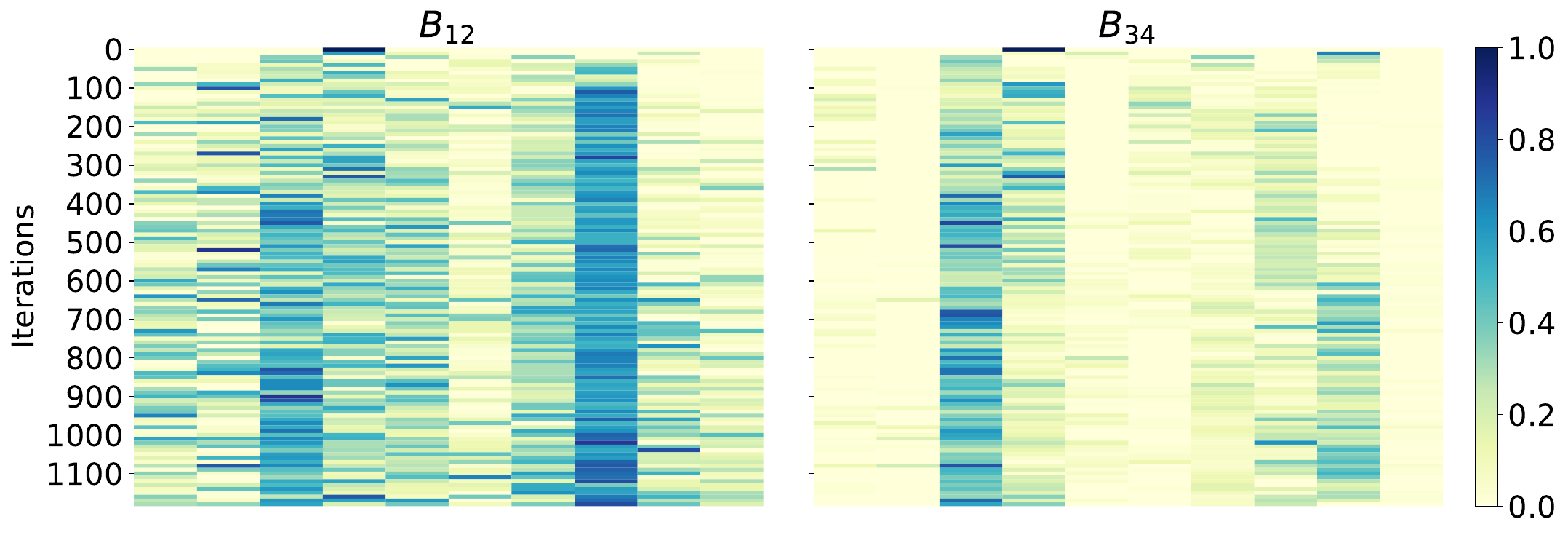}
        \caption{Recall of ResNet50.}
        \label{fig:learimagenet10Recall_resnet50}
        \end{subfigure}
        
    \begin{subfigure}[b]{0.5\textwidth}
        \centering
        \includegraphics[width = \textwidth]{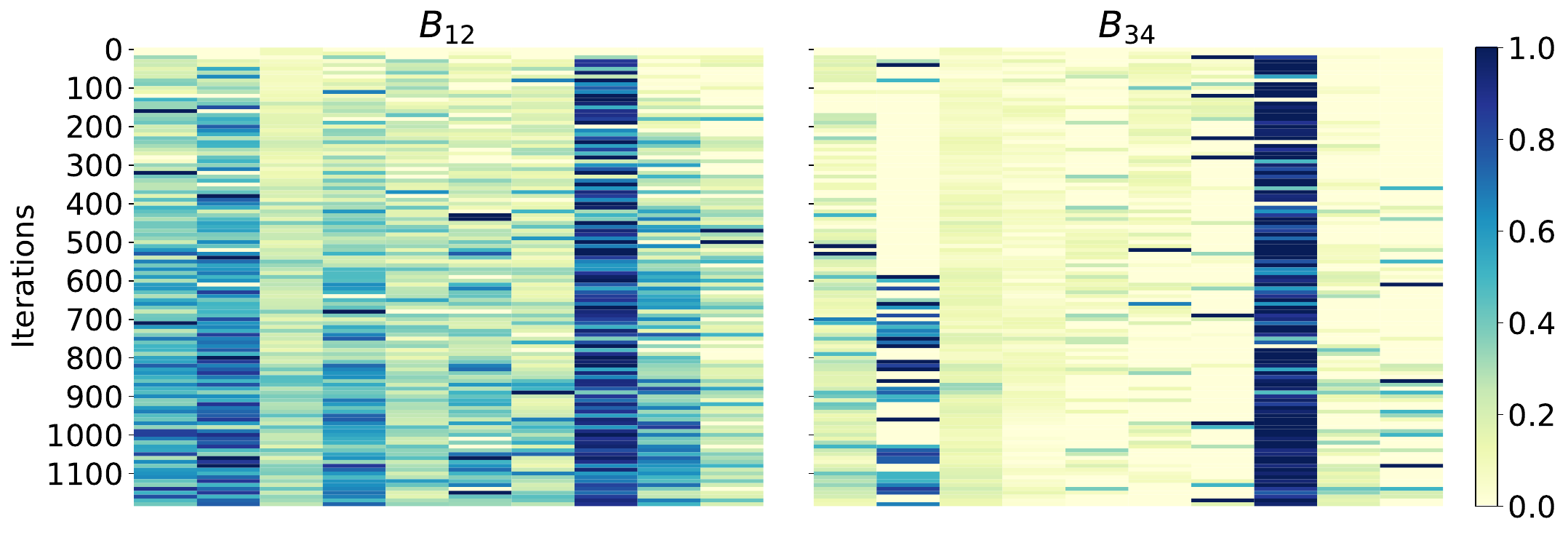}
        \caption{Precision of VGG16.}
        \label{fig:learimagenet10Precision_vgg16}
        \end{subfigure}
    \begin{subfigure}[b]{0.5\textwidth}
        \centering
        \includegraphics[width = \textwidth]{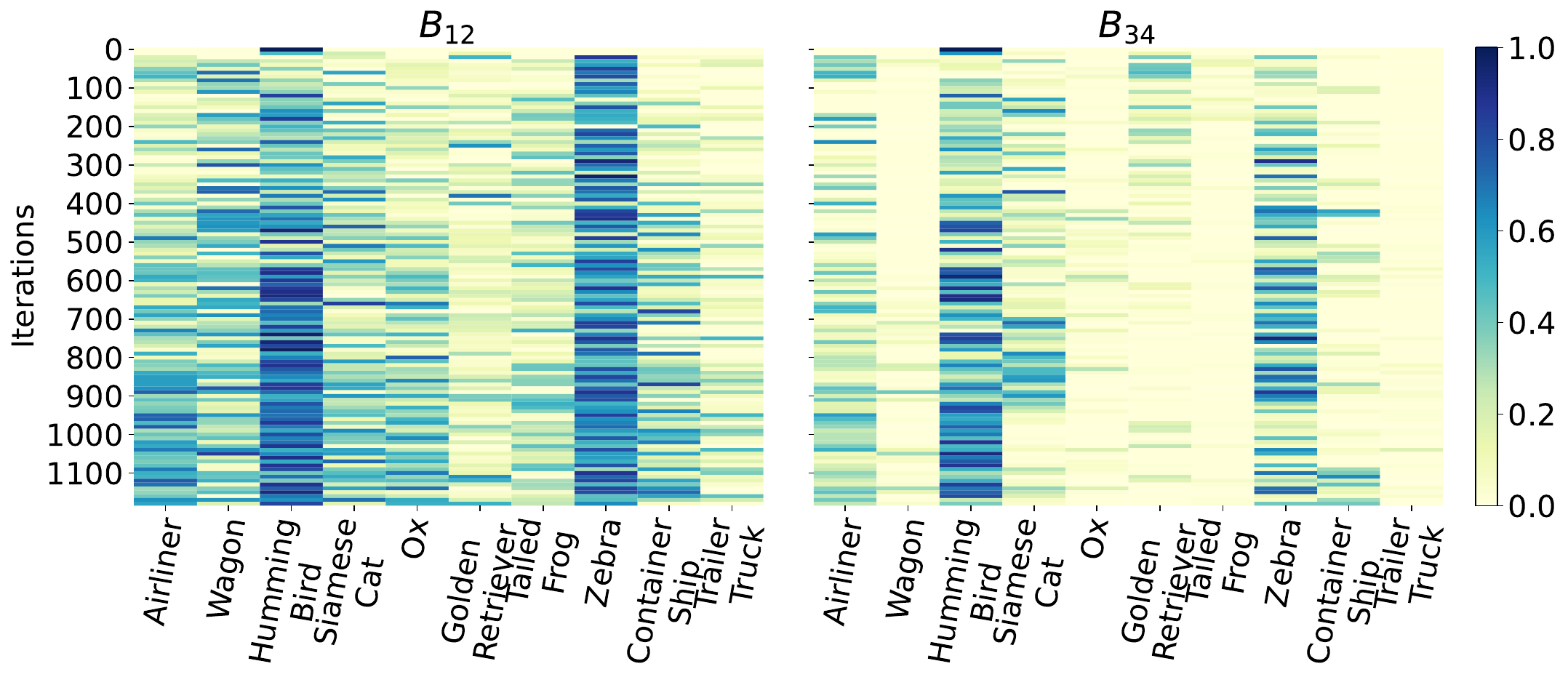}
        \caption{Recall of VGG16.}
        \label{fig:learimagenet10Recall_vgg16}
        \end{subfigure}
     
    \caption{\small{Precision and recall rates of ResNet50 and VGG16 trained on ImageNet-10 for the first $1200$ iterations.}}
    \label{fig:result_imagenet10extra}
\end{figure}

\subsection{ImageNet-10}

\paragraph{ADCS.}
\label{section:adcs}
The ADCS of other classes in ImageNet-10 are shown in~\cref{fig:adcsimagenet}. The classes have different frequency characteristics, which might be applied as discriminative features by NNs for classification. For instance, class `siamese cat' has more energy in low-frequency compared to other classes, which is in line with the observation that the models use more low frequencies to classify the samples of `siamese cat' from the top-$5\%$ DFM. Further, when using SIN (replacing textures while emphasizing shapes) to augment training data, ResNet18 learns a shape-bias frequency shortcut for it, showing the importance to analyze class-wise frequency characteristics of training data in image classification.  Differently, the class `container ship' has more energy on the frequencies whose spatial representations are horizontal and vertical lines. The ADCS of class `trailer truck' shares similar characteristics to that of class `container ship', but it does not have extremely low energy on high-frequency. Similar to the ADCS of class `humming bird', class `ox' has high energy in many high frequencies, though not as high as that of `humming bird'. For other classes without obvious frequency differences, it is difficult to interpret the frequency utilization of the NNs, and thus we compute their DFMs.

\paragraph{Precision and recall.}
We show the precision and recall of ResNet50 and VGG16 computed on the low-passed and high-passed test sets of ImageNet-10 (not the original test set), during the first $1200$ iterations of training in~\cref{fig:result_imagenet10extra}. The models achieve generally higher precision and recall in the classes `humming bird' and `zebra'. This indicates that these classes have special characteristics that are easily used for classification by the models at the early training stages. The observations are in line with the learning behavior of ResNet18 trained on ImageNet-10 that we highlighted in the main paper, confirming that the bias of classification models is indeed driven by data characteristics, being low- or high-frequency components in the images according to the simplicity to solve the optimization problem.

\begin{figure*}[!t]
    \centering
    \begin{subfigure}{\textwidth}
         \centering
         \includegraphics[width=0.8\textwidth]{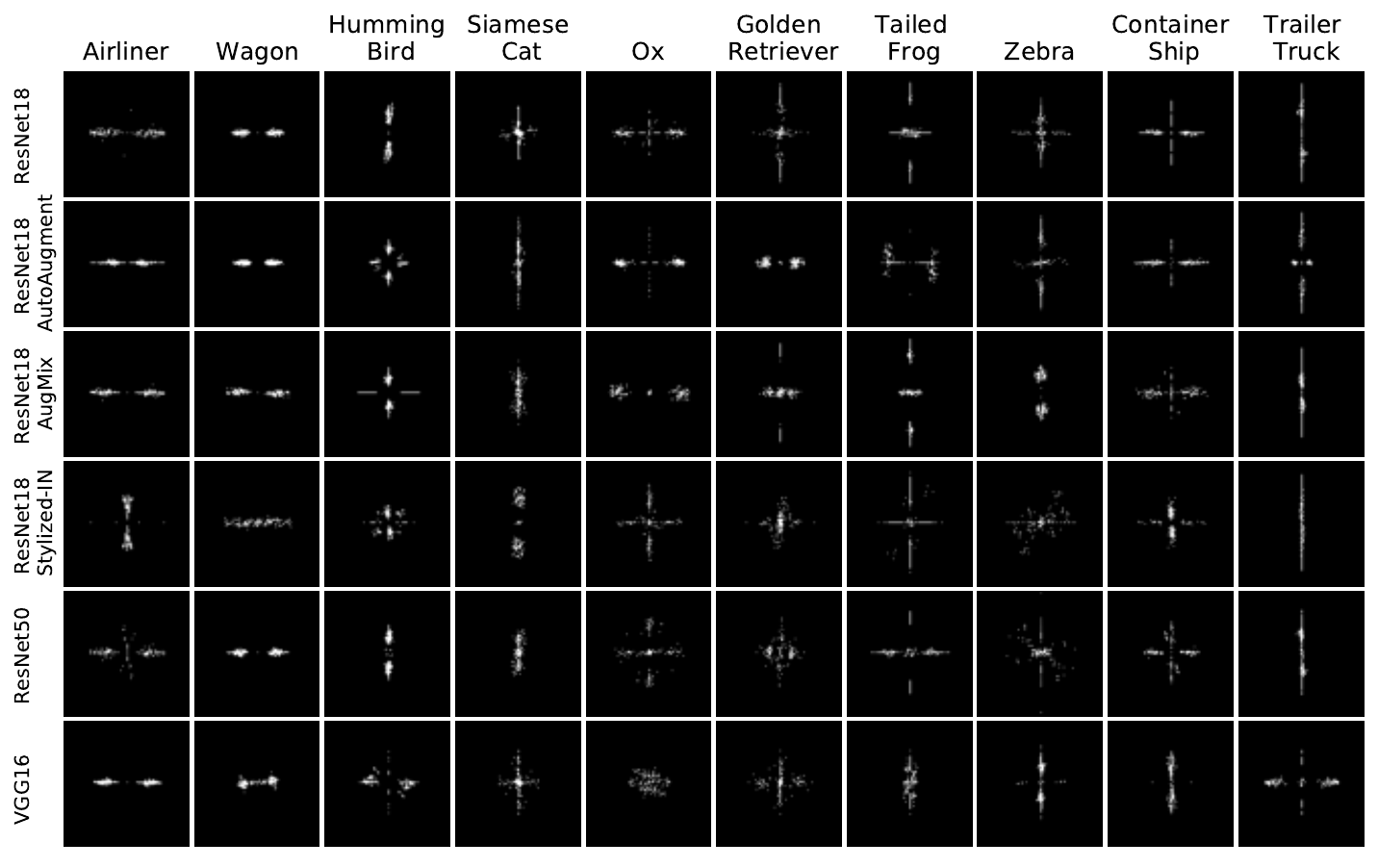}
         \caption{Top-$1\%$}
         \label{fig:masks_rank01}
     \end{subfigure}
    \begin{subfigure}{\textwidth}
         \centering
         \includegraphics[width=0.8\textwidth]{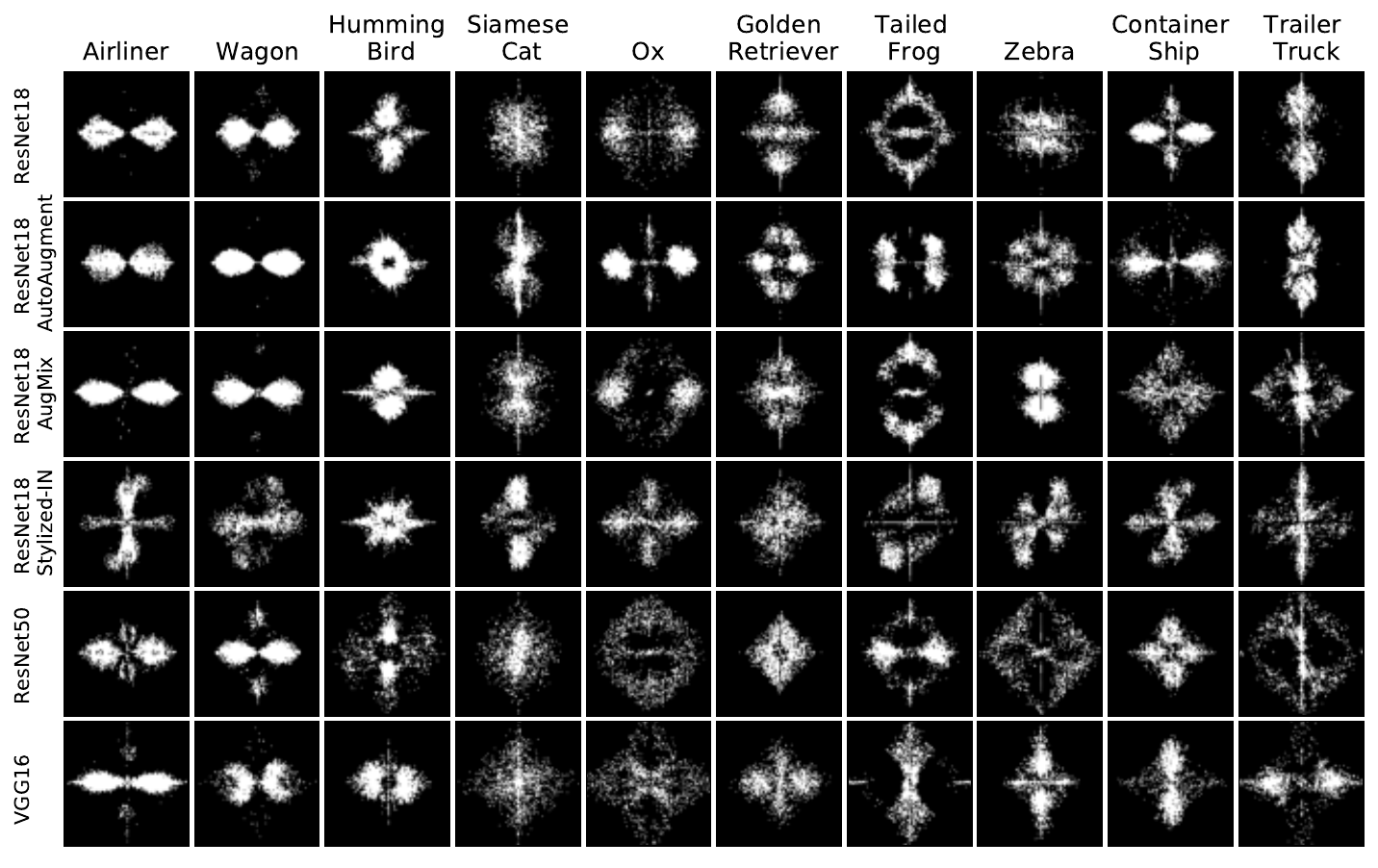}
         \caption{Top-$10\%$}
         \label{fig:masks_rank1}
     \end{subfigure}
    \caption{Dominant frequency maps of ResNet18 (with AutoAugment/AugMix), ResNet50 and VGG16. The maps show the (a) top-$1\%$ and (b) top-$10\%$ dominant frequencies of each class in ImageNet-10.}
    \label{fig:DFMimagenet}
\end{figure*}

\paragraph{Top-$1\%$ and top-$10\%$ DFMs.}

\begin{table*}[!t]
\scriptsize
\addtolength{\tabcolsep}{-5pt}
  \caption{TPRs and FPRs on the top-$1\%$ DFM-filtered versions of ImageNet-10 (w/ df).}
  \label{tab:noFrqs1}
  \centering
  \renewcommand{\arraystretch}{0.95}
  \begin{tabular}{p{2.1cm}c@{\hspace{16\tabcolsep}}c@{\hspace{8\tabcolsep}}c@{\hspace{8\tabcolsep}}c@{\hspace{8\tabcolsep}}c@{\hspace{8\tabcolsep}}c@{\hspace{8\tabcolsep}}c@{\hspace{8\tabcolsep}}c@{\hspace{8\tabcolsep}}c@{\hspace{8\tabcolsep}}c@{\hspace{8\tabcolsep}}c@{\hspace{8\tabcolsep}}c}
    \toprule
    \multicolumn{12}{c}{\bfseries ImageNet-10} \\ \midrule
      \bfseries Model &  & \bfseries airliner  & \bfseries wagon  & \bfseries humming bird & \bfseries siamese cat & \bfseries ox & \bfseries golden retriever & \bfseries tailed frog & \bfseries zebra & \bfseries container ship & \bfseries trailer truck \\
    \midrule
   \bfseries ResNet18   & $TPR$  & 0.08   & 0    & 0      & \textbf{0.84}   & 0.02   & 0      & 0.24   & 0.38   & \textbf{0.4}    & 0.24   \\
    & $FPR$ &0.0067 & 0    & 0      & \textbf{0.1133} & 0.0156 & 0.0089 & 0.0133 & 0.0822 & \textbf{0.1467} & 0.0622 \\
    
    \midrule
    \bfseries ResNet18+AutoAug &$TPR$  &0      & 0    & 0      & 0.06   & 0      & 0      & 0      & 0.08   & 0.14   & 0.04   \\
     & $FPR$ &0      & 0    & 0      & 0.0022 & 0      & 0      & 0.0022 & 0.0756 & 0.1067 & 0.0356 \\
    \midrule
    \bfseries ResNet18+AugMix   & $TPR$  &0.02   & 0      & 0      & 0      & \textbf{0.34 }  & 0.1    & 0.06   & 0    & \textbf{0.94}   & 0.1    \\
     & $FPR$ &  0.0044 & 0      & 0.0044 & 0      & \textbf{0.2133 }& 0.0089 & 0      & 0    & \textbf{0.64}   & 0.04   \\
    \midrule
    \bfseries ResNet18+SIN   & $TPR$  & 0.4 &  0  &   0.22 & \textbf{0.88}    &  \textbf{0.74} &  \textbf{0.72}   &  0  &   0  &  0 &  0.04 \\
     & $FPR$ & 0.1489  &   0.0044    & 0.2444 &   \textbf{0.42}   & \textbf{0.4022} & \textbf{0.4889} &   0.0022   &  0.0022  &  0.0044  &  0.0311 \\
    
    \midrule
    \bfseries ResNet50   & $TPR$  &\textbf{0.34}   & 0      & 0      & 0.12   & 0      & 0.2    & 0      & 0.12   & 0.2    & 0      \\
    & $FPR$ &\textbf{0.1 }   & 0      & 0.0333 & 0.0133 & 0      & 0.04   & 0.0044 & 0.0489 & 0.0556 & 0.0111 \\
    
    \midrule
    \bfseries VGG16 & $TPR$  &0.02   & 0      & 0      & \textbf{0.64}   & \textbf{0.76}   & 0.04   & 0.02   & 0.04   & 0.06   & 0.3    \\
    & $FPR$ &0.0067 & 0      & 0      & 0.0978 & \textbf{0.3556} & 0.0311 & 0      & 0.0422 & 0.06   & 0.1422 \\
    \bottomrule
  \end{tabular}
\end{table*}

\begin{table*}[!t]
\scriptsize
\addtolength{\tabcolsep}{-5pt}
  \caption{TPRs and FPRs on the top-$10\%$ DFM-filtered versions of ImageNet-10 (w/ df).}
  \label{tab:noFrqs10}
  \centering
  \renewcommand{\arraystretch}{0.95}
  \begin{tabular}{p{2.1cm}c@{\hspace{16\tabcolsep}}c@{\hspace{8\tabcolsep}}c@{\hspace{8\tabcolsep}}c@{\hspace{8\tabcolsep}}c@{\hspace{8\tabcolsep}}c@{\hspace{8\tabcolsep}}c@{\hspace{8\tabcolsep}}c@{\hspace{8\tabcolsep}}c@{\hspace{8\tabcolsep}}c@{\hspace{8\tabcolsep}}c@{\hspace{8\tabcolsep}}c}
    \toprule
    \multicolumn{12}{c}{\bfseries ImageNet-10} \\ \midrule
      \bfseries Model &  & \bfseries airliner  & \bfseries wagon  & \bfseries humming bird & \bfseries siamese cat & \bfseries ox & \bfseries golden retriever & \bfseries tailed frog & \bfseries zebra & \bfseries container ship & \bfseries trailer truck \\
    \midrule
   \bfseries ResNet18   & $TPR$  & 0.2    & 0      & 0.62   & 0.92   & 0.06   & 0.16   & 0.12   & \textbf{0.9}    & \textbf{0.84  } & 0.02 \\
    & $FPR$ & 0.0067 & 0      & 0.0378 & 0.0556 & 0.0356 & 0.0156 & 0      & \textbf{0.1156} & \textbf{0.2311} & 0    \\
    
    \midrule
    \bfseries ResNet18+AutoAug & $TPR$  &0    & 0      & 0.22   & \textbf{0.66}   & 0.2    & 0.18   & 0      & \textbf{0.64}   & 0.02   & 0.02   \\
     & $FPR$ &0    & 0      & 0.0067 & \textbf{0.1267} & 0.1067 & 0.0089 & 0      & 0.0289 & 0.0089 & 0.0022 \\
    \midrule
    \bfseries ResNet18+AugMix   & $TPR$  &0.38   & 0      & 0.4    & 0.84  & 0.42   & 0.5    & 0.02   & 0.68   & \textbf{0.9}    & 0.64   \\
    & $FPR$ &0.0356 & 0      & 0.0089 & 0.06   & 0.1556 & 0.0156 & 0      & 0.0022 &\textbf{ 0.1978} & 0.0311 \\
    \midrule
    \bfseries ResNet18+SIN & $TPR$  &  0.12  &   0.04   & 0.6 &  \textbf{0.88}  &  \textbf{0.94}  &  0.62  &   0.06    & 0.66   & 0.08  &  0.12  \\
     & $FPR$ &  0.0089   &    0.0067  & 0.02 & \textbf{0.1044} & \textbf{0.3867} & 0.0933& 0.0022 & 0.0489 & 0.0667 &0 \\
    \midrule
    \bfseries ResNet50   & $TPR$  & 0.44   & 0   & 0.04   & 0.72   & 0      & 0.42   & 0      & 0.12   & \textbf{0.88}   & 0.1    \\
    & $FPR$ &0.0733 & 0   & 0.0044 & 0.0378 & 0.0133 & 0.0311 & 0      & 0.04   & \textbf{0.2356} & 0.0178 \\
    \midrule
    \bfseries VGG16 & $TPR$  & 0.4    & 0      & 0.5    & 0.8    & 0.1    & 0.42   & 0.04 & 0.68   & \textbf{0.82}   & 0.22   \\
     & $FPR$ &0.0422 & 0      & 0.0311 & 0.0467 & 0.1133 & 0.0267 & 0    & 0.0378 & \textbf{0.14}   & 0.0378 \\

    \bottomrule
  \end{tabular}
\end{table*}


We show the top-$1\%$ and top-$10\%$ DFMs of each class for models trained on ImageNet-10 in~\cref{fig:masks_rank01,fig:masks_rank1}.
We observe from the top-$1\%$ DFMs that NNs take the frequencies whose spatial representations are horizontal and vertical lines as the most dominant frequencies since the removal of them results in high loss increment. 
 From the top-$10\%$ DFM in~\cref{fig:masks_rank1}, we observe the frequency utilization of NNs varies slightly across different architectures but shares similar patterns.

\paragraph{Results on ImageNet-10 DFM-filtered versions}

The classification results of models tested on ImageNet-10 DFM-filtered versions, with only the top-$1\%$ and top-$10\%$ dominant frequencies retained, are shown in~\cref{tab:noFrqs1,tab:noFrqs10}.

If a model uses $1\%$ of frequencies and can achieve correct classification for most of the test samples, then it may not extract deep semantic information from the data and be subject to a shortcut learned during training. From~\cref{tab:noFrqs1}, we observe that using only $1\%$ of frequencies, ResNet18+AugMix predicts correctly $94\%$ of the samples of class `container ship' with FPR = 0.64, indicating a learned frequency shortcut and a strong bias towards a small set of frequencies. Interestingly, we observe VGG16, using only $1\%$ of frequencies, learns a frequency shortcut for class `ox', which has TPR = 0.76 and FPR = 0.35. ResNet18+SIN uses frequency shortcuts for classes `siamese cat', `ox', and `golden retriever', observed from the high values of TPR and FPR.

By increasing the number of dominant frequencies considered in the input test images, as expected, all models achieve generally better performance for most of the classes, compared to that on top-1\% DFM-filtered test sets.  From the results of models using the top-$10\%$ dominant frequencies for classification, we can, however, identify similar frequency shortcuts (to the identified frequency shortcuts using the top-$5\%$ dominant frequencies) from the~\cref{tab:noFrqs10}. For instance, models other than ResNet18+AutoAug have high TPRs and FPRs for class `container ship', indicating learned frequency shortcuts. For class `zebra', ResNet18 can predict 90\% of the samples, with FPR = 0.1156, indicating another learned frequency shortcut. Moreover, ResNet18+SIN learns a frequency shortcut for class `ox', while it is less biased to class `siamese cat' with more frequencies provided (lower FPR compared to that of the model tested on the corresponding top-1\% DFM-filtered test set).
The identification of learned frequency shortcuts can be  automatized by choosing the top-$x\%$ ranked frequency and setting thresholds (to the values of TPR and FPR) to evaluate the presence of shortcuts when testing the models on DFM-filtered test sets. 


%% file: imgs/sc_test_new.tex
\begingroup%
  \makeatletter%
  \providecommand\color[2][]{%
    \errmessage{(Inkscape) Color is used for the text in Inkscape, but the package 'color.sty' is not loaded}%
    \renewcommand\color[2][]{}%
  }%
  \providecommand\transparent[1]{%
    \errmessage{(Inkscape) Transparency is used (non-zero) for the text in Inkscape, but the package 'transparent.sty' is not loaded}%
    \renewcommand\transparent[1]{}%
  }%
  \providecommand\rotatebox[2]{#2}%
  \newcommand*\fsize{\dimexpr\f@size pt\relax}%
  \newcommand*\lineheight[1]{\fontsize{\fsize}{#1\fsize}\selectfont}%
  \ifx\svgwidth\undefined%
    \setlength{\unitlength}{450.52008057bp}%
    \ifx\svgscale\undefined%
      \relax%
    \else%
      \setlength{\unitlength}{\unitlength * \real{\svgscale}}%
    \fi%
  \else%
    \setlength{\unitlength}{\svgwidth}%
  \fi%
  \global\let\svgwidth\undefined%
  \global\let\svgscale\undefined%
  \makeatother%
  \begin{picture}(1,1.1337624)%
    \lineheight{1}%
    \setlength\tabcolsep{0pt}%
    \put(0,0){\includegraphics[width=\unitlength,page=1]{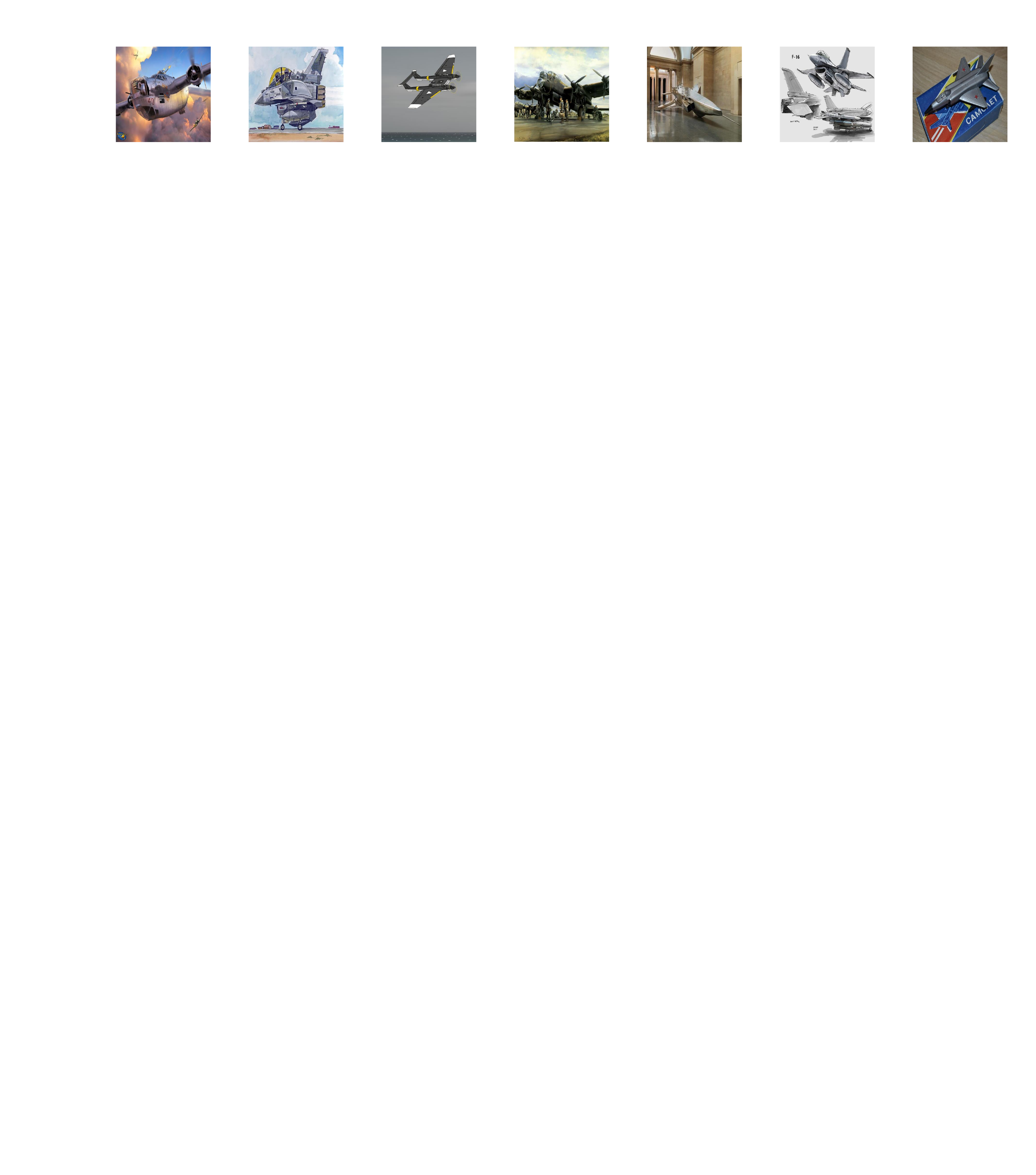}}%
    \put(0.14299006,1.10119414){\color[rgb]{0,0,0}\makebox(0,0)[lt]{\lineheight{1.25}\smash{\begin{tabular}[t]{l}Art \end{tabular}}}}%
    \put(0.2447141,1.10119414){\color[rgb]{0,0,0}\makebox(0,0)[lt]{\lineheight{1.25}\smash{\begin{tabular}[t]{l}Cartoon\end{tabular}}}}%
    \put(0.36108739,1.10119414){\color[rgb]{0,0,0}\makebox(0,0)[lt]{\lineheight{1.25}\smash{\begin{tabular}[t]{l}Deviantart\end{tabular}}}}%
    \put(0.50384482,1.10119414){\color[rgb]{0,0,0}\makebox(0,0)[lt]{\lineheight{1.25}\smash{\begin{tabular}[t]{l}Painting\end{tabular}}}}%
    \put(0.62805645,1.10119414){\color[rgb]{0,0,0}\makebox(0,0)[lt]{\lineheight{1.25}\smash{\begin{tabular}[t]{l}Sculpture\end{tabular}}}}%
    \put(0.76747952,1.10119414){\color[rgb]{0,0,0}\makebox(0,0)[lt]{\lineheight{1.25}\smash{\begin{tabular}[t]{l}Sketch\end{tabular}}}}%
    \put(0.90998494,1.10119414){\color[rgb]{0,0,0}\makebox(0,0)[lt]{\lineheight{1.25}\smash{\begin{tabular}[t]{l}Toy\end{tabular}}}}%
    \put(0,0){\includegraphics[width=\unitlength,page=2]{imgs/sc_test_new_ig.pdf}}%
    \put(0.02459677,1.05103647){\color[rgb]{0,0,0}\makebox(0,0)[lt]{\lineheight{1.25}\smash{\begin{tabular}[t]{l}Military \end{tabular}}}}%
    \put(0.02704714,1.02312084){\color[rgb]{0,0,0}\makebox(0,0)[lt]{\lineheight{1.25}\smash{\begin{tabular}[t]{l}aircraft\end{tabular}}}}%
    \put(0.04550248,0.93196209){\color[rgb]{0,0,0}\makebox(0,0)[lt]{\lineheight{1.25}\smash{\begin{tabular}[t]{l}Car\end{tabular}}}}%
    \put(0.02282878,0.82641257){\color[rgb]{0,0,0}\makebox(0,0)[lt]{\lineheight{1.25}\smash{\begin{tabular}[t]{l}Lorikeet\end{tabular}}}}%
    \put(0.01558623,0.71650124){\color[rgb]{0,0,0}\makebox(0,0)[lt]{\lineheight{1.25}\smash{\begin{tabular}[t]{l}Tabby cat \end{tabular}}}}%
    \put(0.02227047,0.6065899){\color[rgb]{0,0,0}\makebox(0,0)[lt]{\lineheight{1.25}\smash{\begin{tabular}[t]{l}Holstein\end{tabular}}}}%
    \put(0.01882754,0.51304023){\color[rgb]{0,0,0}\makebox(0,0)[lt]{\lineheight{1.25}\smash{\begin{tabular}[t]{l}Labrador \end{tabular}}}}%
    \put(0.01999069,0.4851246){\color[rgb]{0,0,0}\makebox(0,0)[lt]{\lineheight{1.25}\smash{\begin{tabular}[t]{l}retriever\end{tabular}}}}%
    \put(0.01819169,0.3867737){\color[rgb]{0,0,0}\makebox(0,0)[lt]{\lineheight{1.25}\smash{\begin{tabular}[t]{l}Tree frog\end{tabular}}}}%
    \put(0.03356079,0.27684298){\color[rgb]{0,0,0}\makebox(0,0)[lt]{\lineheight{1.25}\smash{\begin{tabular}[t]{l}Horse\end{tabular}}}}%
    \put(0.0283964,0.1717458){\color[rgb]{0,0,0}\makebox(0,0)[lt]{\lineheight{1.25}\smash{\begin{tabular}[t]{l}Fishing \end{tabular}}}}%
    \put(0.03273883,0.14383017){\color[rgb]{0,0,0}\makebox(0,0)[lt]{\lineheight{1.25}\smash{\begin{tabular}[t]{l}vessel\end{tabular}}}}%
    \put(0.01657878,0.04787536){\color[rgb]{0,0,0}\makebox(0,0)[lt]{\lineheight{1.25}\smash{\begin{tabular}[t]{l}Fire truck\end{tabular}}}}%
    \put(0,0){\includegraphics[width=\unitlength,page=3]{imgs/sc_test_new_ig.pdf}}%
  \end{picture}%
\endgroup%